\documentclass{article} 
\usepackage{iclr2025_conference,times}

\usepackage{amsmath,amsfonts,bm}

\def\eqref#1{equation~\ref{#1}}

\def\1{\bm{1}}

\DeclareMathAlphabet{\mathsfit}{\encodingdefault}{\sfdefault}{m}{sl}
\SetMathAlphabet{\mathsfit}{bold}{\encodingdefault}{\sfdefault}{bx}{n}

\usepackage{hyperref}
\usepackage{url}
\usepackage{graphicx}
\usepackage{booktabs}
\usepackage{subcaption}
\usepackage{multirow}
\usepackage{array}
\usepackage{bm}
\usepackage{makecell}
\usepackage{enumitem}
\usepackage{hyperref}
\usepackage{minitoc}

\newcommand{\mname}{WorldSimBench}
\newcommand{\mc}{\textcolor[HTML]{80C2AE}{Open-Ended Embodied Environment}}
\newcommand{\abmc}{\textcolor[HTML]{80C2AE}{OE}}
\newcommand{\ad}{\textcolor[HTML]{C280B5}{Autonomous Driving}}
\newcommand{\abad}{\textcolor[HTML]{C280B5}{AD}}
\newcommand{\arm}{\textcolor[HTML]{C2AE80}{Robot Manipulation}}
\newcommand{\abarm}{\textcolor[HTML]{C2AE80}{RM}}
\newcommand{\upstream}{Explicit Perceptual Evaluation}
\newcommand{\downstream}{Implicit Manipulative Evaluation}

\newcommand{\pretextmodel}{Predictive Text Model}
\newcommand{\preimagemodel}{Predictive Image Model}
\newcommand{\prevideomodel}{Predictive Video Model}
\newcommand{\preacvideomodel}{Predictive Actionable Video Model}

\newcommand{\dataset}{HF-Embodied Dataset}
\newcommand{\datalen}{35,701}
\newcommand{\evaluator}{Human Preference Evaluator}
\newcommand{\abevaluator}{HPE}
\newcommand{\eg}{\textit{e.g.}}
\newcommand{\ie}{\textit{i.e.}}

\title{WorldSimBench: Towards Video Generation  Models as World Simulators}

\author{
Yiran Qin\textsuperscript{1,2}\footnotemark[1]~, 
Zhelun Shi\textsuperscript{3}\footnotemark[1]~, 
Jiwen Yu\textsuperscript{4}, 
Xijun Wang\textsuperscript{2}, 
Enshen Zhou\textsuperscript{3}, 
Lijun Li\textsuperscript{2},  \\
\textbf{~Zhenfei Yin\textsuperscript{2}\footnotemark[3], 
Xihui Liu\textsuperscript{4}, 
Lu Sheng\textsuperscript{3}, 
Jing Shao\textsuperscript{2}\footnotemark[2]~,
Lei Bai\textsuperscript{2}\footnotemark[2],
Wanli Ouyang\textsuperscript{2},
Ruimao Zhang\textsuperscript{1}\footnotemark[2]}\\
\\
\small $^{1}$The Chinese University of Hong Kong, Shenzhen
\small$^{2}$ Shanghai Artificial Intelligence Laboratory\\
\small$^{3}$Beihang University ~~
\small$^{4}$The University of Hong Kong~~~\\
\\
~~~~~~~~~~~~~~~~~~~~~~~\textcolor[HTML]{CC96A6}{Project Page: \href{https://iranqin.github.io/WorldSimBench.github.io}{https://iranqin.github.io/WorldSimBench.github.io}}
}

\iclrfinalcopy 
\begin{document}

\maketitle

\let\thefootnote\relax\footnotetext{$^*$ Equal contribution\hspace{3pt} \hspace{5pt}$^\dagger$ Corresponding author\hspace{5pt} $^\ddagger$ Project lead
}

\begin{abstract}

Recent advancements in predictive models have demonstrated exceptional capabilities in predicting the future state of objects and scenes.
However, the lack of categorization based on inherent characteristics continues to hinder the progress of predictive model development.
Additionally, existing benchmarks are unable to effectively evaluate higher-capability, highly embodied predictive models from an embodied perspective.
In this work, we classify the functionalities of predictive models into a hierarchy and take the first step in evaluating World Simulators by proposing a dual evaluation framework called \mname.
\mname ~includes \textbf{\upstream} and \textbf{\downstream}, encompassing human preference assessments from the visual perspective and action-level evaluations in embodied tasks, covering three representative embodied scenarios: \mc, \ad, and \arm.
In the \upstream, we introduce the \dataset, a video assessment dataset based on fine-grained human feedback, which we use to train a \evaluator{} that aligns with human perception and explicitly assesses the visual fidelity of World Simulators.
In the \downstream, we assess the video-action consistency of World Simulators by evaluating whether the generated situation-aware video can be accurately translated into the correct control signals in dynamic environments.
Our comprehensive evaluation offers key insights that can drive further innovation in video generation models, positioning World Simulators as a pivotal advancement toward embodied artificial intelligence.

\end{abstract}

\begin{figure}[h]
\begin{center}
\includegraphics[width=1\linewidth]{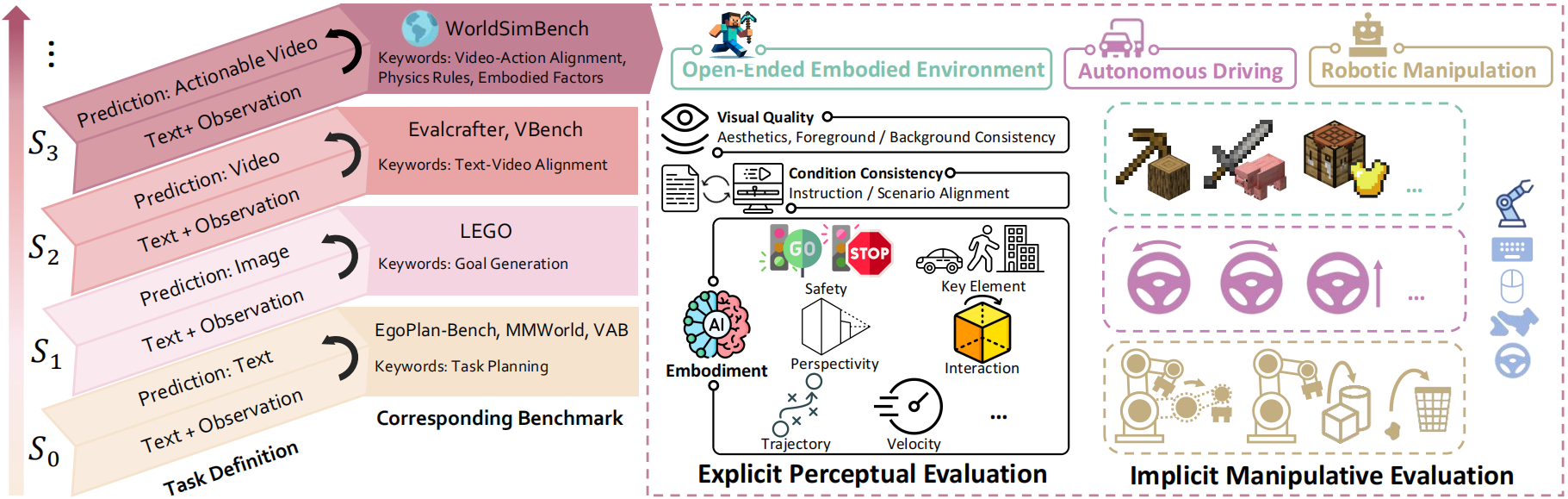}
\end{center}
\caption{\textbf{Overview of the hierarchical capabilities of the Predictive Models.} Models at higher stages demonstrate more advanced capabilities. We take the initial step in evaluating Predictive Generative Models up to the $S_3$ stage, known as World Simulators, by introducing a parallel evaluation framework, \mname. 
\mname~assesses the models both \upstream{} and \downstream, focusing on video generation and action transformation across three critical embodied scenarios. 
} 
\label{fig:motivation}
\end{figure}

\section{Introduction}
Before taking action, humans make predictions based on their objectives and observations of the current environment. 
These predictions manifest in various forms, \eg, textual planning, visual imagination of future scene changes, or even subconscious planning at the action level.
%
%
With the development of generative models, agents driven by these models are exhibiting predictive capabilities that enable them to complete embodied tasks by making human-like predictions, \eg{}, high-level planning~\citep{driess2023palm,li2024manipllm}, image-based guidance~\citep{lai2023lego,black2023zero}, or future video prediction to drive actions~\citep{du2023video,du2024learning}). We refer to these models as \textbf{Predictive Models}.
Recently, these models have been widely applied across various domains spanning from developing agents to solve inference tasks to leveraging predictions for driving robots to perform specific actions.

Nevertheless, the rich application scenarios and diverse model designs make predictive models a broad family.
However, without categorizing them based on their inherent characteristics, the advancement of predictive model development remains limited.
This leads to our first question: \textit{Can we establish a reasonable hierarchical system for Predictive Models based on their degree of embodiment?}
With a well-defined categorization, we can better target the evaluation of Predictive Models from different perspectives of embodiment, ensuring that their strengths and weaknesses are adequately assessed.

In the literature, existing evaluations have typically focused on task planning capabilities by assessing text outputs or evaluating visual outputs from an aesthetic perspective. 
However, such approaches significantly limit the evaluation of highly embodied Predictive Models, as embodied scenarios are more concerned with physical properties (\eg, perspective consistency, object breakability), which these methods fail to effectively assess. 
This brings us to our second question: \textit{Can we conduct a more detailed evaluation of highly embodied Predictive Models from an embodied perspective?}

To answer the first question, we categorize the functionalities of Predictive Models into a hierarchy from $S_0$ to $S_3$, defined by the model's capabilities and level of embodiment, accompanied by corresponding evaluation benchmarks as illustrated in Fig.~\ref{fig:motivation}.
Models are classified based on the degree of embodiment in their output modalities.
From lower to higher stages, the models are capable of generating: text, images, videos, and actionable videos (\ie, the videos that can be translated into actions).
It is worth noting that Predictive Models at $S_3$ capable of generating actionable videos integrate robust 3D scene understanding and physical rule priors to provide precise guidance for generating executable actions. These models are closely aligned with the recently proposed concept of World Simulators~\citep{yang2023learning}.

To answer the second question, we review the related benchmarks, as listed in Tab.~\ref{tab: benchmark_comparison}. 
Evaluations on models in $S_0$ that generate text primarily focus on assessing task planning capabilities, while $S_1$ and $S_2$ assessments on visual output measure aesthetic quality through feature similarity analyses with ground truth data. 
With clearly defined evaluation dimensions and extensive annotated datasets, both types of assessments can be effectively conducted.
However, evaluating World Simulators introduces complexities due to the intricate physical definitions involved. 
Additionally, conventional evaluation methods are inadequate for assessing the actionablilty of the generated videos, as there is no definite ground truth for actionable videos towards completing a specific embodied task.
These factors pose significant challenges to the evaluation of World Simulators.

\begin{table}[t]
\centering
\tiny
\caption{\textbf{Comparisons between existing Predictive Model benchmarks.} Interactive Environment refers to the interaction with the simulation environment during the prediction phase. Task-Level Interaction denotes that each task interacts once, whereas Action-Level Interaction represents the frequency of interactions that occur through the generation of actions for control purposes.}
\vspace{-3mm}
\begin{tabular}{@{}lccccccc@{}}
\toprule
Benchmark & Input Modality & Output Modality& Based Method & Stage & Interactive Env. & Evaluation Strategy  \\
\midrule
AgentBench~\citep{liu2023agentbench} & Text & Text&LLM & $S_0$ & Task-Level & Human Judgement   \\
EgoPlan-Bench~\citep{chen2023egoplan} & Text \& Images & Text & MLLM & $S_0$ & N/A & Multi-choice   \\
MMWorld~\citep{he2024mmworld} & Text \& Images & Text & MLLM & $S_0$ & N/A & GPT Judgement  \\
VAB~\citep{liu2024visualagentbench} & Text \& Images & Text & MLLM & $S_0$ & Task-Level & Human Judgement  \\
LEGO~\citep{lai2023lego} & Text \& Images & Image &IGM& $S_1$ & Task-Level & Feature Similarity  \\
VBench~\citep{huang2024vbench} & Text & Video &VGM& $S_2$ & N/A & Feature Similarity  \\
EvalCrafter~\citep{liu2024evalcrafter} & Text \& Images & Video &VGM & $S_2$ & N/A & Feature Similarity  \\
\midrule
\multirow{2}{*}{\centering \mname} & \multirow{2}{*}{\centering Text \& Images} & \multirow{2}{*}{Actionable Video} & \multirow{2}{*}{VGM}& \multirow{2}{*}{\centering $S_3$} & \multirow{2}{*}{\centering Action-Level} & \evaluator{} \\

& & & & & & Embodied Metric \\
\bottomrule
\end{tabular}
\label{tab: benchmark_comparison}
\vspace{-5mm}
\end{table}

We argue that an evaluation aligned with human perception could provide a more intuitive and accurate reflection of the characteristics of the synthesized videos, including their adherence to physical rules.
Besides, the actionability can be assessed through a closed-loop manner in simulations deployed with a unified video-to-action policy network.
Considering these aspects, we take the very first step in evaluating World Simulators by proposing a dual evaluation framework called \mname.
As shown in Fig.~\ref{fig:motivation}, \mname{}~assesses World Simulators through two complementary approaches: \textbf{\upstream}, which focuses on the Visual Quality, Condition consistency, and Embodiment of the generated content, and \textbf{\downstream}, which measures the World Simulator's performance through the conversion of video into control signals. 
We present three representative embodied scenarios: \mc{} (\abmc), \ad{} (\abad), and \arm{} (\abarm), to thoroughly evaluate the capability of World Simulators in generating and representing scenario-specific attributes.

In the \upstream, we first define evaluation criteria which is used to construct a comprehensive set of prompts specific to each scenario. 
The prompt lists are then used by various video generation models to produce a large number of video clips.
Following extensive human feedback and annotation, these video clips are compiled into the HF-Embodied dataset which consists of a total of \datalen{} tuples with multi-dimensional scores and fine-grained human feedback.
Additionally, we train \evaluator{}, using the HF-Embodied dataset to assess World Simulators at the perceptual level, offering a robust evaluation of both their visual fidelity and contextual accuracy.
For the \downstream, we deploy three simulation environments for the three embodied scenarios respectively. These environments are used to collect data and train inverse dynamic or goal-based video-to-action models capable of mapping future videos to actions.
In each of these embodied scenarios, the World Simulator is tasked with generating situation-aware videos in real-time, based on current observations and provided text instructions.
These generated videos are then converted into actions using the pre-trained video-to-action models. 
The effectiveness of the World Simulator is implicitly evaluated by measuring the performance of the tasks, using relevant metrics to reflect the quality and accuracy of the generated video.

In summary, the main contributions are as follows:
(1)We categorize the functionalities of Predictive Models into a hierarchy, defined by the model's capabilities and level of embodiment, to advance research and development in the field and take the very first step in evaluating World Simulators.
(2)We propose a dual evaluation framework called \mname, through \upstream{} and \downstream{}, we conducted a comprehensive evaluation of the World Simulator's capabilities from an embodied perspective, focusing on both the visual and action levels.
(3)We conducted extensive testing across multiple models and performed a thorough analysis of the experimental results. Our findings highlight the strengths and limitations of current World Simulators and provide actionable insights for improving future video generation models.
(4)We developed \dataset{}, which includes fine-grained human feedback across three scenarios and 20 dimensions, with a total of \datalen entries. This dataset, containing both human ratings and the reasons behind them, not only enables the evaluation of World Simulators but also provides broader applications~(\eg, alignment) for future video generation models.

\section{Related Work}
\label{related_word}

\textbf{Predictive Models.}
Predictive models are capable of generating process representations that map the current state to future states by incorporating current state representations and control over future trends. 
\pretextmodel, built on LLMs~\citep{radford2019language,touvron2023llama,chiang2023vicuna} and MLLMs~\citep{openai2023gpt4,geminiteam2023gemini,llava,yin2023lamm}, generate future predictions in the text modality by accepting current state representations and text instructions. 
These models have demonstrated impressive performance in high-level planning tasks for embodied agents~\citep{driess2023palm,li2024manipllm,qin2024mp5,chen2024rh20t,zhang2024ad, lu2024gpt4geminibeyondassessing}. 
Similarly, image generation models~\citep{brooks2023instructpix2pix,fu2023guiding} as \preimagemodel~\citep{lai2023lego,black2023zero,zhou2024minedreamer} can produce future goal images, showcasing strong capabilities during the decision-making phase of embodied agents. 
\prevideomodel~\citep{du2024learning,du2023video}, based on video generation models~\citep{janner2022planning}, have made some progress in embodied control. 
However, due to limitations in data or models, the generated videos often lack essential physical representations and logical consistency, restricting their applicability to fixed scenarios and single tasks.

With the advancement of diffusion transformer~\citep{peebles2023scalable} and the extensive utilization of large-scale internet video datasets~\citep{bain2021frozen,ebert2021bridge,goyal2017something,grauman2022ego4d}, certain \preacvideomodel~\citep{yang2023learning} models, also known as World Simulators, have achieved more precise representations of physical laws and 3D environments.

\textbf{Evaluation of Predictive Models.} With the advancement of predictive models, research has also expanded to evaluate the capabilities of models at different stages. 
\cite{liu2023agentbench,chen2023egoplan,shi2024assessmentmultimodallargelanguage,liu2024visualagentbench} conducted text-level and task completion evaluations for \pretextmodel ~at the $S_0$ stage. 
\cite{lai2023lego} performed score-based evaluations from an aesthetic perspective for \preimagemodel ~at the $S_1$ stage. 
\cite{huang2024vbench,liu2024evalcrafter} also assessed the aesthetic quality of videos generated by \prevideomodel ~at the $S_2$ stage. 
We take the first step in evaluating World Simulators through an embodied perspective.

\section{Predictive Model Category Definition}

In this section, we concretely categorize predictive models based on the model's capabilities and level of embodiment. 
The detailed categorization stage of Fig.~\ref{fig:motivation} is illustrated below,

\textbullet ~\textbf{Stage} $S_0$: At this stage, predictive models can generate corresponding predictions based on instructions and observations but are limited to textual modality. Benchmarks at this stage conduct text-level and task-completion evaluations through output text planning.

\textbullet ~\textbf{Stage} $S_1$: At this stage, predictive models can generate visual predictions based on instructions and observations, but without incorporating temporal information. Benchmarks at this stage conduct aesthetic evaluation for generated images.

\textbullet ~\textbf{Stage} $S_2$: At this stage, predictive models can generate corresponding video predictions based on both instructions and observations. Yet, due to limited model capabilities, the evaluation at this level focuses solely on the aesthetic quality of the generated outputs.

\textbullet ~\textbf{Stage} $S_3$: At this stage, predictive models can generate corresponding video predictions based on instructions and observations, with the predicted video content adhering to physical rules and aligning with the executed actions. These models are known as \textbf{World Simulators}~\citep{ha2018world,yang2023learning}, and \mname ~is a benchmark specifically designed to evaluate these World Simulators.

The rapidly evolving field of World Simulators offers exciting opportunities for advancing Artificial General Intelligence, with significant potential to enhance human productivity and creativity, especially in embodied intelligence. Therefore, conducting a comprehensive embodied evaluation of World Simulators is crucial.

\section{WorldSimBench Construction}

\begin{figure}[h]
\begin{center}
\includegraphics[width=1\linewidth]{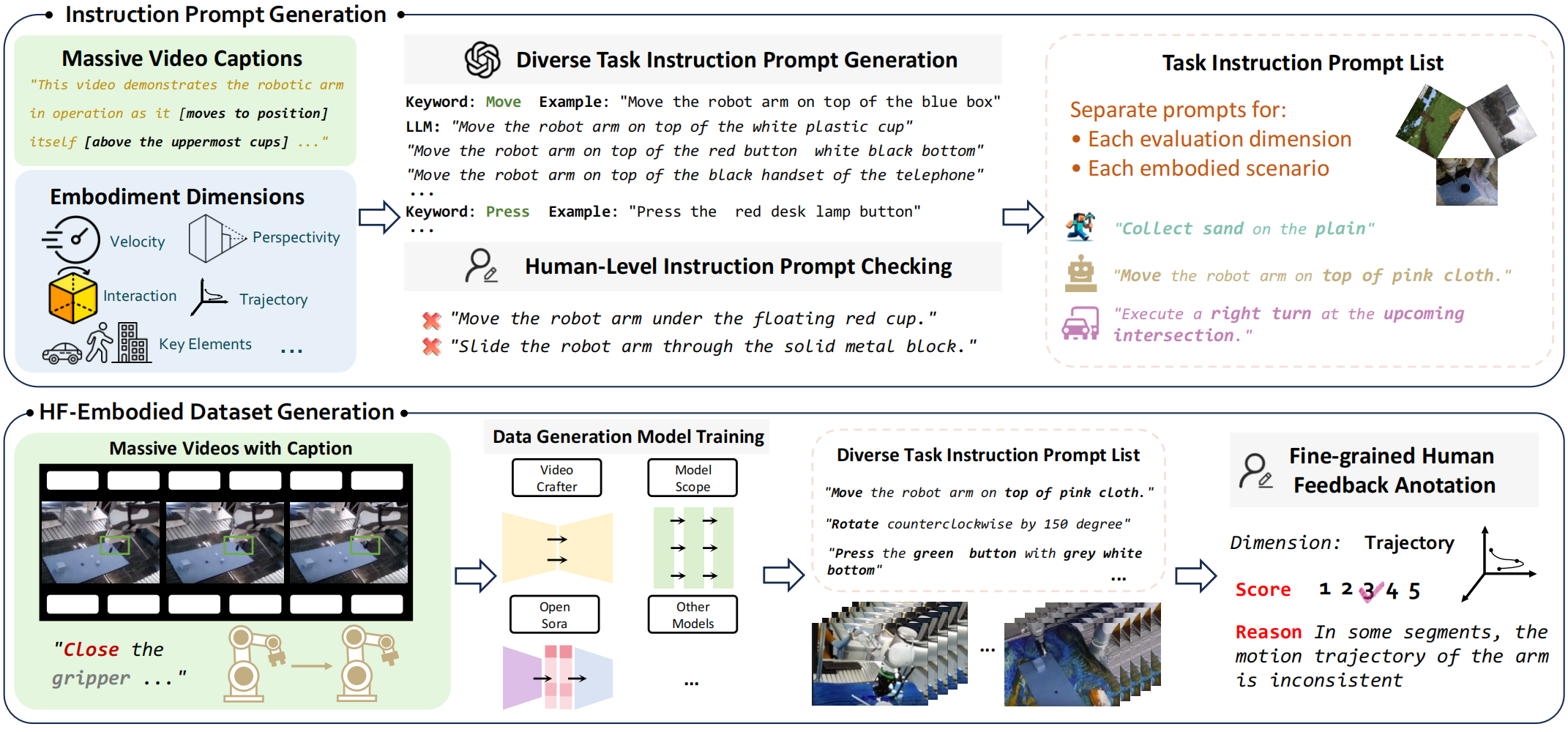}
\end{center}
\vspace{-3mm}
\caption{Overview of \textbf{\upstream}.
(Top)~\textbf{Instruction Prompt Generation.} We use a large collection of video captions from the internet and our predefined embodied evaluation dimensions. These are expanded using GPT and manually verified to create a corresponding Task Instruction Prompt List for data generation and evaluation.
(Bottom)~\textbf{HF-Embodied Dataset Generation.} Massive internet-sourced embodied videos with captions are used to train data generation models. Fine-grained Human Feedback Annotation is then applied to the embodied videos according to the corresponding Task Instruction Prompt List, covering multiple embodied dimensions.} 
\label{fig:Perceptual_Evaluation}
\vspace{-5mm}
\end{figure}

WorldSimBench evaluates the embodied capabilities of World Simulators across two distinct levels. The \textbf{\upstream} assesses the simulators based on human-perceived quality across different embodied scenarios, while the \textbf{\downstream} implicitly evaluates the simulators' capabilities by converting the generated videos into control signals and observing their performance in various closed-loop embodied tasks.

The evaluation of World Simulators encompasses three critical embodied scenarios: \mc{} (\abmc), \ad{} (\abad), and \arm{} (\abarm). 
Minecraft serves as a popular testbed for \abmc, providing a challenging platform for agents to handle complex, unstructured tasks. 
In the context of \abad, especially in outdoor settings, ensuring the stability and robustness of the agent’s actions is crucial, making it an essential domain for assessing a World Simulator’s capability in dynamic and uncertain environments. 
\abarm, a core task in embodied intelligence, demands precise and adaptive control, testing the world simulator’s ability to generate actionable predictions that align with physical interactions. 
Together, these scenarios provide a comprehensive benchmark for evaluating the effectiveness of World Simulators across a range of real-world tasks.

\subsection{\upstream}

In \upstream, we propose Hierarchical Evaluation Dimensions, based on which we build a video assessment dataset annotated through fine-grained human feedback, named \dataset. The dataset is constructed based on three key resources, each corresponding to a specific embodied scenario: a curated dataset of Minecraft videos from the internet for \abmc~\citep{baker2022video}, real-world driving data for \abad~\citep{caesar2020nuscenes}, and real-world robot manipulation videos annotated with text instructions for \abarm~\citep{chen2024rh20t}. 
Using \dataset, we train a \evaluator{} to perform perceptual evaluations of World Simulators.

\subsubsection{Hierarchical Evaluation Dimension}

We develop a hierarchical evaluation dimension checklist for the three embodied scenarios, as illustrated in Tab.~\ref{tab:hierarchical_evaluation_dimension}, which can be categorized into three main aspects: \textbf{Visual Quality}, \textbf{Condition Consistency}, and \textbf{Embodiment}. 
(1)~Visual Quality primarily assesses the overall quality of video generation, including Aesthetics, Background and Foreground Consistency. 
(2)~Condition Consistency focuses on the alignment with the input instruction. For tasks in \abmc{} that involve distinct scenarios, we additionally define Scenario Alignment to assess the alignment to the specific scenarios outlined in the instruction.
(3)~Embodiment has different definitions depending on the scenario. 
As all tasks require movement along a certain trajectory, we uniformly define Trajectory to evaluate the rationality of object movement in the video (\eg{}, whether a robotic arm avoids obstacles during motion). In \abad{} and \abarm{}, we define Perspectivity to assess whether the video exhibits a clear sense of depth. In \abmc{} and \abarm{}, we define Embodied Interaction to evaluate the plausibility of interactions with objects. We also define Velocity in \abmc{} to determine whether speed varies appropriately across different environments (\eg{}, slower movement in water). In \abad, we define Key Element to evaluate the rendering quality and consistency of crucial embodied elements, \eg, pedestrians. We also introduce Safety in \abad{} to assess whether the embodied actions comply with traffic rules.
More details in Sup.~\ref{sec: taxonomy_upstream}.


\begin{table}[t]
    \centering
    \footnotesize
    \caption{\textbf{Hierarchical Evaluation Dimension.} The dimensions are categorized into three main aspects: Visual Quality for evaluating the overall quality, Condition Consistency for evaluating the alignment to the input instruction, and Embodiment for evaluating embodied related factors like physical rules.}
    \vspace{-3mm}
    \renewcommand{\arraystretch}{1.2}
    \resizebox{1.0\textwidth}{!}{
    \begin{tabular}{l|ccc}
        \toprule
        Embodied Scenarios & Visual Quality & Condition Consistency & Embodiment \\
        \midrule
        \mc{} (\abmc) & \makecell{Background Consistency (BC) \\ Foreground Consistency (FC)} & \makecell{Instruction Alignment (IA) \\ Scenario Alignment (SA)} & \makecell{Velocity (VC) \\ Trajectory (TJ) \\ Embodied Interaction (EI)}  \\
        \hline
        \ad{} (\abad) & \makecell{Aesthetics (AE)} & \makecell{Instruction Alignment (IA)} & \makecell{Perspectivity (PV) \\ Trajectory (TJ) \\ Key Element (KE) \\ Safety (SF)}\\
        \hline
        \arm{} (\abarm) & \makecell{Aesthetics (AE) \\ Background Consistency (BC) \\ Foreground Consistency (FC)} & Instruction Alignment (IA) & \makecell{Perspectivity (PV) \\ Trajectory (TJ) \\ Embodied Interaction (EI) } \\
        \bottomrule
    \end{tabular}
    }
    \label{tab:hierarchical_evaluation_dimension}
    \vspace{-5mm}
\end{table}

\subsubsection{Instruction Prompt Generation}  
Using the Hierarchical Evaluation Dimension and massive video captions from the key resources, we create a foundational but comprehensive prompt list. We utilize the knowledge of LLMs, \ie{} ChatGPT, to extend the meta-prompts across a wide range. After manual screening for relevance, diversity, and data distribution, we compile the Task Instruction Prompt List, which separates prompts for each content-embodied scenario and each evaluation dimension, as shown in Fig.~\ref{fig:Perceptual_Evaluation}.

\subsubsection{\dataset{} Generation}

\textbf{Data Preparation.} 
We select multiple video generation models and train them using a large corpus of videos and corresponding captions from the key resources. 
Due to the capabilities of the open-source video generation model, we conduct targeted training for each of the three distinct embodied scenarios individually, thereby developing several data generation models for different embodied scenarios. 
These models are then used to produce a substantial amount of instruction-following embodied videos, based on the corresponding captions, and the initial image condition where applicable (first frame conditioned text-to-video to generate situation-aware videos).


\textbf{Human Annotation.}
We use human annotation to label the generated videos. Based on the Hierarchical Evaluation Dimension, we establish specific annotation guidelines and numerous in-conttext examples for the annotators. For each dimension, annotators are instructed to score the video solely based on its performance within that particular dimension and provide corresponding reasoning. For instance in \abarm{}, as illustrated in Fig.~\ref{fig:Perceptual_Evaluation}, under the dimension of Trajectory, annotators are required to evaluate the video exclusively on the generation quality of the motion trajectory. They are instructed not to consider other elements (\eg{}, the rendering quality of the robot arm) or other dimensions (\eg{}, consistency with instructions). Additionally, annotators are asked to provide fine-grained feedback on any deficiencies, \eg, ``inconsistent trajectory". 
As a result, we construct the \dataset{}, which consists of a total of \datalen{} tuples, each comprising a video, text instruction, multi-dimensional scores, and the potential reasons. More details in Sup.~\ref{subsec: data_analysis}.

\subsubsection{\evaluator} 
The objective is to develop a video scoring model that assesses videos across multiple dimensions aligning with human perception. The model takes a generated video and a prompt as input and outputs a score ranging from 1 to $n$ ($n$ is defined specifically for each embodied scenario). The prompt includes both the video generation instructions and an explanation of the evaluation criteria.
Leveraging the strong video understanding capabilities of multimodal large language models, we fine-tune Flash-VStream~\citep{zhang2024flashvstreammemorybasedrealtimeunderstanding}, a VideoLLM, aligning it with human perception on \dataset{}. 
Only LoRA~\citep{hu2021loralowrankadaptationlarge} parameters are trained. 
This enables the model to effectively grasp the evaluation metrics for embodied tasks and produce accurate scores, while maintaining its video perception and reasoning ability. We prove the effectiveness and generalizability of our \evaluator{} in Sec.~\ref{sec:evaluator_effectiveness}.

\subsubsection{Evaluation Metrics.}
\label{sec: upstream_eval_metrics}
The evaluation of a video generation model is based on the scores assigned by the evaluator across various dimensions. For each dimension, the video generation model generates videos guided by several carefully selected instructions sourced from Task Instruction Prompt List that are strongly aligned with the specific evaluation criteria, \eg{}, ``explore on the beach'' for Embodied Scenario in \abmc. The final metric for each model is computed as the average score across all dimensions. The evaluated dimensions for each embodied scenario are listed in Tab.~\ref{tab:hierarchical_evaluation_dimension}.

\begin{figure}[t]
\begin{center}
\includegraphics[width=1\linewidth]{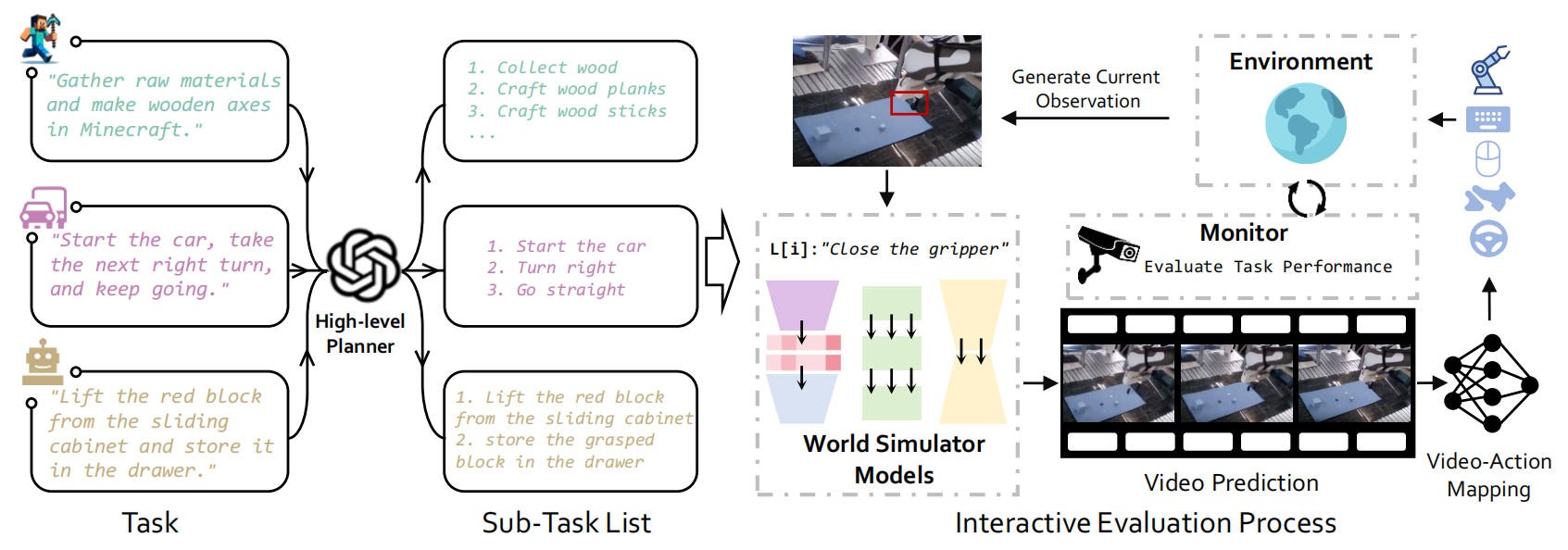}
\end{center}
\vspace{-3mm}
\caption{Overview of \textbf{\downstream}.
Embodied tasks in different scenarios are decomposed into executable sub-tasks. The video generation model generates corresponding predicted videos based on the current instructions and real-time observations. Using a pre-trained IDM or a goal-based policy, the agent executes the generated sequence of actions. After a fixed timestep, the predicted video is refreshed by sampling again from the video generation model, and this process repeats. Finally, the success rates of various embodied tasks are obtained through monitors in the simulation environment.} 
\label{fig:Action_Level_Evaluation}
\vspace{-6mm}
\end{figure}

\subsection{\downstream}
The Implicit Manipulative Evaluation assesses the capabilities of World Simulators across various embodied scenarios by treating the simulator as a low-level decision maker for situational contexts. Using pre-trained video-to-action models, we implicitly evaluate the performance of the World Simulators by observing their effectiveness in closed-loop embodied task tests.

\subsubsection{Simulation construction} 
The Implicit Manipulative Evaluation is conducted using the following three simulation platforms, for specific settings, please refer to the Supplementary Material.

\abmc{} We employ MineRL as the Minecraft simulator, with the observation space limited to RGB images and the action space confined to keyboard and mouse controls. We adopt the Steve-1 benchmarks~\citep{lifshitz2024steve}, with task descriptions \eg, "chop a tree." 

\abad{} We conduct standard closed-loop evaluations using the CARLA~\citep{dosovitskiy2017carla} simulator on the LangAuto Benchmark~\citep{shao2024lmdrive}. Task descriptions include instructions like "do not deviate from this route." 

\abarm{} We employ CALVIN~\citep{mees2022calvin} as the robot manipulation simulator, using only RGB images for the observation space and limiting the action space to the 7-DOF (degrees of freedom) of the robot arm. Task descriptions include commands \eg, "pull the handle to open the drawer."

\subsubsection{Embodied task evaluation} 
\textbf{Evaluation Pipeline.} As illustrated in Fig.~\ref{fig:Action_Level_Evaluation}, we first leverage existing or custom-trained video-to-action models as intermediaries between the World Simulator and the agent performing closed-loop tasks, for the selected benchmarks across three simulation environments. This approach enables the transformation of the predicted future videos from the World Simulator into executable control signals in real-time, thereby indirectly evaluating the World Simulator's capability through the successful completion of embodied tasks. The evaluation process is tailored to the specific nature of the models under consideration, establishing distinct protocols for closed-loop task evaluation. We fine-tune the models on simulation datasets tailored to each task. These datasets, derived from the three aforementioned benchmarks, include task instructions and corresponding videos, ensuring the models are well-adapted to the specific embodied scenarios. Finally, the evaluated World Simulator is integrated with the video-to-action model to jointly form an embodied agent that performs the given tasks. The agent’s performance across various tasks serves as a direct measure of the World Simulator's effectiveness.

\textbf{Evaluation Metrics.} 
\label{down_Metrics}
In \abmc, we track the MineRL~\citep{guss2019minerl} environment state to calculate metrics \eg, travel distance and early-game item collection. Travel distance is the agent's maximum horizontal displacement (X-Z plane) from the spawn point, while dig depth is its maximum vertical displacement (Y axis). We record the maximum number of logs, seeds, and dirt items in the agent's inventory during the episode. In \abad, we employ eight widely used evaluation metrics in Carla~\citep{dosovitskiy2017carla}, including Route Completion (RC), Infraction Score (IS), Driving Score (DS), Vehicle Collisions (VC), Pedestrian Collisions (PC), Layout Collisions (LC), Red Light Violations (RV), and Offroad Infractions (OI). In \abarm, we evaluate the video generation model in the CALVIN~\citep{mees2022calvin} setting (train on A, B, C → test on D) by running 20 trials and calculating the average success rate.

\begin{table}[t]
    \footnotesize
    \centering
    \caption{\textbf{The overall performance comparison between \evaluator{} and GPT-4o.} \abevaluator{} indicates \evaluator. HPE@Lavie means that HPE is trained on videos except those generated by Lavie. The validation is conducted on videos generated by Laive under zero-shot setting.}
    \vspace{-3mm}
    \resizebox{\textwidth}{!}{
    \begin{tabular}{l|cc|cc|cc}
        \toprule
        Embodied Scenario        & GPT-4o    &\abevaluator   &GPT-4o@OpenSora    &\abevaluator@OpenSora  &GPT-4o@Lavie   &\abevaluator@Lavie \\
        \midrule 
        \abmc@Acc(\textuparrow)  & 72.8      & \bf 89.4      & 66.5              & \bf 71.6              & 78.5          & \bf 87.9          \\
        \abad@PLCC(\textuparrow) & 0.28      & \bf 0.60      & 0.03              & \bf0.34               & -0.04         & \bf 0.49          \\
        \abarm@PLCC(\textuparrow)& 0.07      & \bf 0.43      & -0.06             & \bf 0.47              & 0.17          & \bf 0.44          \\
        \bottomrule
    \end{tabular}}
    \label{tab: ab_comp_eval_gpt}
    \vspace{-4mm}
\end{table}

\section{Experiments}
\label{sec:exp}

\subsection{Experimental Setup}
\label{sec:exp_setup}

We evaluate 8 popular video generation models, including Open-Sora-Plan(T2V)~\citep{pku_yuan_lab_and_tuzhan_ai_etc_2024_10948109}, Lavie~\citep{wang2023lavie}, ModelScope~\citep{wang2023modelscope}, OpenSora~\citep{opensora}, AnimateDiff~\citep{guo2023animatediff}, Open-Sora-Plan(TI2V)~\citep{pku_yuan_lab_and_tuzhan_ai_etc_2024_10948109}, Dynamicrafter~\citep{xing2023dynamicrafter}, EasyAnimate~\citep{xu2024easyanimate} through both \upstream{} and \downstream, across three distinct scenarios: \mc{} (\abmc), \ad{} (\abad), and \arm{} (\abarm). All models finetuned on specific datasets corresponding to three embodied scenarios in \upstream{} and \downstream{}. Detailed information on the datasets, training, and testing configurations can be found in the Supplementary Material.
 
For \upstream{}, we extract five instructions from the Task Instruction Prompt List for each dimension across the three embodied scenarios, ensuring they strongly align with the specific evaluation criteria, as discussed in Sec.~\ref{sec: upstream_eval_metrics}. 
The selected instruction prompts each model to generate five videos, which are then scored by the \evaluator{} to obtain an average score for the model's performance. 
For the scoring range 1-$n$, $n$ is set 2 for \abmc{}, and set 5 for both \abad{} and \abarm{}. 
We indicate that the generation quality in \abmc ~is perceived as binary from a human perspective, while the other two scenarios exhibit a more diverse range of video quality.

For \downstream{}, we constructed three video-to-action models for embodied simulation environments, following the designs of Steve-1~\citep{lifshitz2024steve}, Susie~\citep{black2023zero}, and LMdrive~\citep{shao2024lmdrive}. For the evaluated models, we used the following datasets for fine-tuning: (1) VPT~\citep{baker2022video} and our own collected videos along with corresponding task descriptions as the training set for the \abmc{}; (2) the full Calvin(ABC\_D)~\citep{mees2022calvin} video dataset and corresponding robot arm control instructions as the training set for \abarm{}; and (3) the full Carla~\citep{dosovitskiy2017carla} video dataset and corresponding autonomous driving navigation commands as the training set for \abad{}. Since the video-to-action model in our \abmc ~setup utilizes a goal-based policy, which interprets the goal from the input video and generates actions based on the current observations and the goal, it allows us to additionally evaluate text-to-video models. 

\begin{figure}[t]
    \begin{center}
    \includegraphics[width=0.9\linewidth]{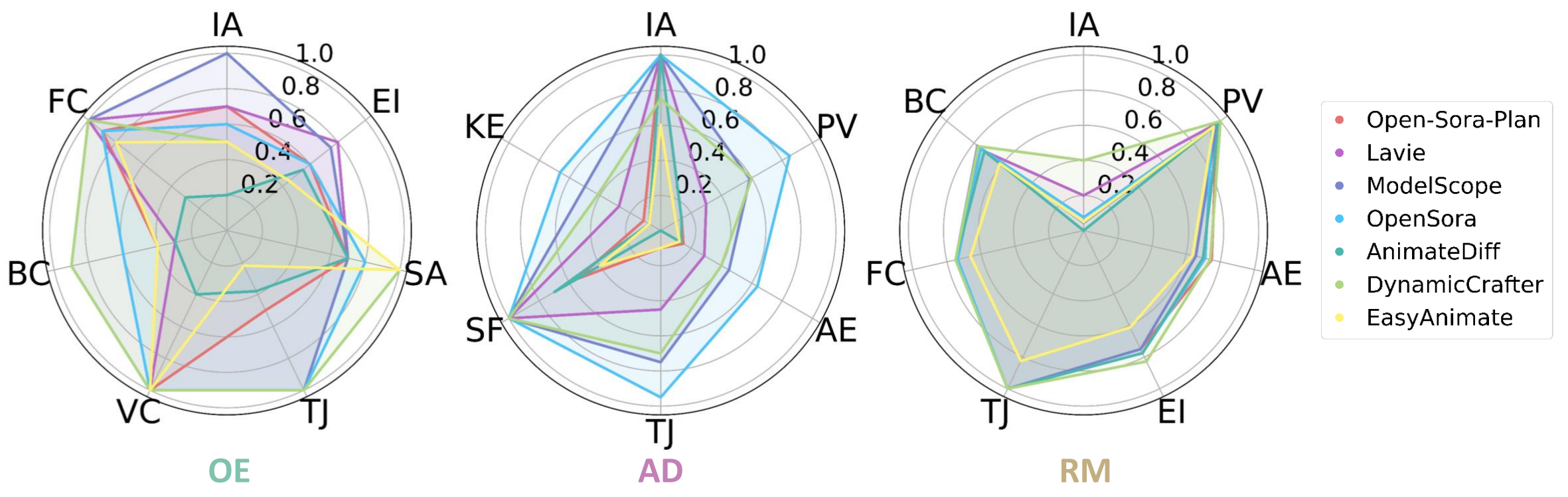}
    \end{center}
    \vspace{-4mm}
    \caption{\textbf{Result of \upstream{} aross three embodied scenarios.} Scores in each embodied scenario are normalized to 0-1. The abbreviations are listed in Tab.~\ref{tab:hierarchical_evaluation_dimension}.}
    \label{fig:EPE_benchmark_result}
\vspace{-4mm}
\end{figure}

\subsection{Experiments on \evaluator}
\label{sec:evaluator_effectiveness}
We demonstrate the strong capabilities and generalization of \evaluator{} by comparing it with GPT-4o~\citep{gpt4o}, showcasing its applicability for \upstream{}, as shown in Tab.~\ref{tab: ab_comp_eval_gpt}. We use accuracy (Acc) in \abmc{} to assess the alignment of the model with human preferences, given the scoring range of 1-2. In contrast, we employ Pearson linear correlation coefficient (PLCC) for \abad{} and \abarm{} as their scores range from 1-5.

After fine-tuning on \dataset, our evaluator consistently surpasses the performance of GPT-4o in terms of alignment with human preferences across all scenarios. Additionally, we conducted zero-shot experiments with two challenging models, \ie{} OpenSora and Lavie. GPT-4o exhibits a negative correlation with human preferences in evaluating OpenSora in \abad{} under zero-shot setting, as well as evaluating Lavie in \abarm{} under zero-shot setting. Our evaluator's zero-shot performance shows a high correlation with human preferences, further demonstrating its robust generalization capabilities. \evaluator{} is suitable for \upstream, and the \dataset{} can be leveraged to train even more aligned models for assessing video generation models towards World Simulators. More details in Sup.~\ref{sup:evaluator}.

\subsection{Design Features and Discussions}
In this section, we discuss the Design features and corresponding observations we draw from our comprehensive evaluation experiments. More details can be found in the Supplementary Material.


\textbf{Human Prefrence with Feedback.} Given the complexity and diversity in the representation of physical rules in videos, even a specific dimension may manifest in various ways (for example, both illogical and discontinuous object motion fall under trajectory-related issues). This makes it challenging to evaluate using score-based models or a single fixed set of evaluation criteria. \mname ~addresses this challenge effectively by employing a human preference scoring mechanism and a fine-grained feedback system. 
Fig.~\ref{fig:EPE_benchmark_result} illustrates the evaluation results of \upstream, more detail analyze could be found in Sup.~\ref{Detailed_Result_of_upstream}.
In \abmc, most models struggle with Embodied Interaction, particularly in generating plausible object deformations, \eg, block shattering, due to the complexity of physical rules. 
In \abad, the variation between models is minimal, with high-performing models excelling across all dimensions. The simpler instructions, like moving forward or turning, lead to high Instruction Alignment, but many generated videos suffer from poor 3D depth (Perspectivity) and fail to depict realistic embodied elements like pedestrians and vehicles, affecting the overall Aesthetic.
In \abarm, models perform uniformly well in static scene depiction, excelling in Perspectivity and Foreground/Background Consistency. However, they struggle with Instruction Alignment, often generating aimless actions. Despite this, the lack of unreasonable trajectories results in relatively high Trajectory scores, though robotic manipulation remains a significant challenge for current models.

\begin{figure}[t]
    \begin{center}
    \includegraphics[width=0.9\linewidth]{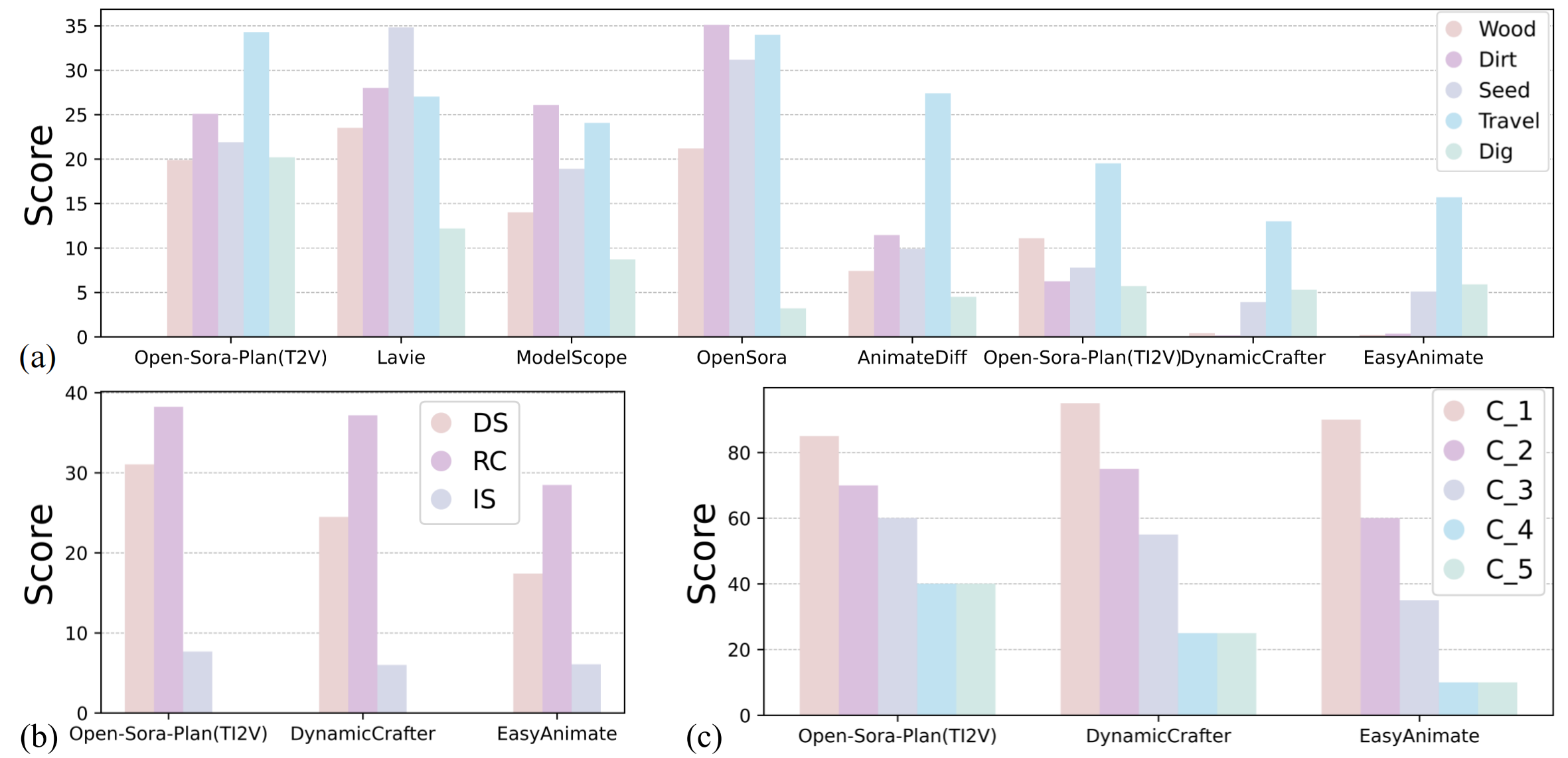}
    \end{center}
    \caption{\textbf{Result of \downstream{} aross three embodied scenarios.}The abbreviations are listed in Sec.~\ref{down_Metrics}.} 
    \label{fig:expfig_downstream}
\vspace{-6mm}
\end{figure}

\textbf{Close-loop Interactive Evaluation.} Given the dynamic nature and real-time requirements of interactive environments, evaluating World Simulators through static benchmarks often fails to capture the full spectrum of their capabilities. Close-loop Interactive Evaluation addresses this by enabling continuous feedback and adaptation, ensuring that the model's predictions and actions evolve in response to the changing environment, thus providing a more accurate and realistic assessment of its performance. 
Fig.~\ref{fig:expfig_downstream} presents the \downstream evaluation results, showing significant variation in the performance of video generation models across different tasks. In the \abmc, video generation models conditioned on the first frame have a significantly lower success rate compared to those without image conditioning. This suggests that models with image conditioning struggle to generate physical laws and 3D scene representations accurately.
Tasks like travel, requiring high-quality trajectories and 3D representation, show the greatest variation in model performance, while simpler tasks like collecting wood see similar performance across models, indicating effective handling of minimal background variation. In the \abad, models with better trajectory(Open-Sora-Plan) generation perform better.
In the \abarm, where background variation is minimal, models perform similarly on simple tasks, but as complexity increases, more robust models achieve higher success rates. Despite some success across scenarios, video generation models still need significant improvements in generating physically consistent content to be reliable for training agents or guiding actions. 

\textbf{Alignment of Physical Rules and Actions.} Ensuring that World Simulators adhere to physical laws while generating predictions is crucial for practical application. The alignment of physical rules and actions is essential as it guarantees that the model's outputs are not only visually plausible but also executable in real-world scenarios. This approach allows for the seamless integration of predicted actions with their physical environment, ensuring reliability and effectiveness in real-world tasks.
Based on our experimental findings, we observe that most conclusions from the \upstream and \downstream evaluations are consistent. Specifically, the visual quality across most dimensions aligns with the results from the closed-loop experiments. \eg, Dynamicrafter, which performs well in trajectory generation in \upstream, also excels in trajectory-focused scenarios like \abad ~and \abarm. However, in other cases—such as the \abmc, which requires more frequent interactions, and long-sequence tasks (4, 5) in \abarm—Dynamicrafter underperforms compared to Open-Sora-Plan. This differs from the \upstream{} results, likely because these tasks demand stable, high-quality video generation for guidance, where Open-Sora-Plan shows higher robustness. Therefore, a comprehensive evaluation of video generation models requires a combination of \upstream{} and \downstream{} assessments to provide the most fair and accurate judgment.
Finally, based on the overall \upstream and \downstream results, we conclude that current video generation models still fail to effectively capture many physical rules, indicating significant improvements are needed before they can function as true World Simulators.

\section{Conclusion}
In this work, we classify the functionalities of predictive models into a hierarchy and take the first step in evaluating World Simulators by proposing a dual evaluation framework called \mname.
We conducted a comprehensive evaluation and analysis of multiple video generation models as World Simulators through both \upstream{} and \downstream{} processes. 
We summarize key findings from the evaluation and hope these insights will inspire and guide future research on World Simulators.

\textbf{Limitations.} Although we evaluate physical rules and 3D content from the perspective of embodied intelligence, the World Simulator can be applied to more scenarios than just robots, and different scenarios have more physical representations, so how to effectively evaluate the World Simulator in other scenarios requires more exploration.

\bibliography{iclr2025_conference}
\bibliographystyle{iclr2025_conference}

\clearpage
\doparttoc
\faketableofcontents
\part{Appendix} 
\parttoc 
\clearpage
\appendix

\section{Taxonomy in \upstream}
\label{sec: taxonomy_upstream}
We outline the evaluation dimensions for each embodied scenario below, along with their corresponding explanations. These explanations are used for detailed human annotation documentation and also serve as the  explanation of the evaluation criteria in instructions for the \evaluator{}.

\subsection{\mc}
\textbf{Visual Quality.} Background Consistency ensures the background remains consistent throughout the video. Foreground Consistency verifies the consistency of the foreground elements.

\textbf{Condition Consistency.} Instruction Alignment assesses whether the video aligns with the provided input instruction. Scenario Alignment checks if the input instruction defines an embodied scenario and whether the video accurately reflects this scenario.

\textbf{Embodiment.} Velocity evaluates if the velocity of the observed object is appropriate. Embodied Interaction evaluates the embodied interaction's appropriateness based on the interaction process and target.
Trajectory evaluates whether the motion trajectory in the video is logical.

\subsection{\ad}
\textbf{Visual Quality.} Aesthetics evaluates whether the composition, color, lighting, and scene in the video align with human aesthetics.

\textbf{Condition Consistency.} Instruction Alignment assesses whether the video aligns with the provided input instruction.

\textbf{Embodiment.} Perspectivity evaluates the video's perspective, specifically assessing the 3D scene relationships. This includes evaluating whether the video has a strong sense of depth and realism (\ie, whether it feels three-dimensional). Additionally, assess the logic of lighting and shadows, including whether the shadow positions are consistent with the light sources. Trajectory evaluates whether the movement and the trajectory of elements in the video is logical. Key Element assesses the generated quality of embodied elements \eg, roads, vehicles, pedestrians, bicycles, lane markings, sidewalks, traffic signs, and traffic lights. Safety evaluates whether the behavior of the vehicles comply with traffic rules. Are there any instances of running red lights, speeding, or driving outside of permissible areas.

\subsection{\arm}
\textbf{Visual Quality.} Aesthetics evaluates whether the composition, color, lighting, and scene in the video align with human aesthetics. Background Consistency ensures the background remains consistent throughout the video, include the manipulation table and the environment. Foreground Consistency verifies the consistency of the foreground elements, including the robotic arm and the object on the manipulation table.

\textbf{Condition Consistency.} Instruction Alignment assesses whether the action of the robot arm in the generated video aligns with the provided input instruction.

\textbf{Embodiment.} Perspectivity evaluates the video's perspective, specifically assessing the 3D scene relationships. This includes evaluating whether the video has a strong sense of depth and realism (\ie., whether it feels three-dimensional). Additionally, assess the logic of lighting and shadows, including whether the shadow positions are consistent with the light sources. Embodied Interaction judges whether the object's shape and posture conform to the rules during the collision of objects and the interaction between the robotic arm and the object. Trajectory evaluates whether the trajectory of the robotic arm is reasonable and in line with human cognition.

\section{Detaild Implementation of \upstream}

\subsection{\dataset{}}
\label{subsec: data_analysis}
\begin{table}[t]
    \centering
    \caption{\textbf{Analysis of \dataset{}.} Samples scored higher than 3 in \abad{} and \abarm{} are considered positive.}
    \resizebox{\textwidth}{!}{
    \begin{tabular}{l|cccccc}  
        \toprule
        Embodied Scenario   & \#instructions& \#videos  & \#dims    & \#actions & \#positive& \#negative\\ 
        \hline
        \mc{}               & 270           & 8401      & 7         & 11        & 121249    & 79965     \\ 
        \ad{}               & 5             & 15870     & 6         & 5         & 56768     & 35044     \\ 
        \arm{}              & 2556          & 11430     & 7         & 26        & 70672     & 9338      \\
        \bottomrule
    \end{tabular}}
    \label{tab: analysis_dataset}
\end{table}

Tab.~\ref{tab: analysis_dataset} provides an analysis of the \dataset{}. In \ad{} scenario, there are only five instructions: move forward, move backward, turn left, turn right, and stop. The other two scenarios include a variety of instructions that combine actions with target objects. Given the diverse instructions, different video generation models generate multiple videos after finetuning on specific datasets. To enhance the \evaluator{} understanding of the autonomous driving context, we also supplement the \abad{} scenario with videos from real-world scenes. Additionally, we list the quantities of positive and negative samples across all dimensions. Samples with human annotated scores of 3 or higher in \abad{} and \abarm{} are considered positive. Leveraging \dataset{} with comprehensive embodied dimensions, we train the \evaluator{} to enable efficient assessment in \upstream{}.

\subsection{Video Generation Model Finetuning}
\label{sup_b2}


\begin{table}[htbp]
    \centering
    \caption{\textbf{Training Frames of Generation Models.}}
    \resizebox{\textwidth}{!}{
    \begin{tabular}{l|ccccccc}  
        \toprule
        Model & Open-Sora-Plan & Lavie & ModelScope & OpenSora & AnimateDiff & DynamicCrafter & EasyAnimate\\ 
        \hline
        Short Videos(frames)    & 16 & 16 & 16 & 16 & 16 & 16 & 16\\
        Long Videos(frames)     & 64 & 48 & 60 & 48 & 64 & 60 & 64\\
        \bottomrule
    
    \end{tabular}}
    \label{tab_training_frames}
\end{table}

We evaluate 8 popular video generation model, including Open-Sora-Plan(T2V)~\citep{pku_yuan_lab_and_tuzhan_ai_etc_2024_10948109}, Lavie~\citep{wang2023lavie}, ModelScope~\citep{wang2023modelscope}, OpenSora~\citep{opensora}, AnimateDiff~\citep{guo2023animatediff}, Open-Sora-Plan(TI2V)~\citep{pku_yuan_lab_and_tuzhan_ai_etc_2024_10948109}, DynamicCrafter~\citep{xing2023dynamicrafter}, EasyAnimate~\citep{xu2024easyanimate} through both \upstream{} and \downstream, across three distinct scenarios: \mc{} (\abmc), \ad{} (\abad), and \arm{} (\abarm). 

In \mc, we use \textbf{OpenAI Contractor Gameplay Dataset}~\citep{baker2022video} which is created by hiring human contractors to play Minecraft and complete tasks like house building.
Keypresses and mouse movements are recorded during gameplay. We apply the same preprocessing steps as VPT, including filtering out null actions. 
Additionally, we create a supplementary dataset for the task "Explore" by generating trajectories using various pre-trained Steve-1 agents. 
The distribution of this dataset is enhanced by randomly switching between models during trajectories, resetting the agent’s memory, and adjusting the agent’s orientation to face new directions at random intervals.
For specific in-game events, \eg, ``\texttt{mine\_block}'', the type of block broken is logged alongside precise timestamps. 
These timestamps allow for accurate progress tracking and are aligned with the completion of event-related instructions.

In \ad, we fine-tune using the nuScenes training set~\citep{caesar2020nuscenes}, and following the approach in Vista~\citep{gao2024vista}, we sample video clips consisting of 25 frames at a frequency of 10 Hz. 
To classify actions into textual commands, we adhere to established conventions in planning and define ego-vehicle commands as ``turn right'', ``turn left'', ``go straight'', and ``stop'', consistent with the definitions in Vista.

In \arm, we use RH20T-P~\citep{chen2024rh20t}, a dataset based on RH20T~\citep{fang2023rh20t} and designed for primitive-level robotic manipulation that features meticulously defined primitive skills and diverse primitive-level spatial knowledge of multiple forms. 
We use each primitive-level robotic manipulation instruction along with the corresponding video as input for training. 
Additionally, since this dataset is designed for downstream tasks in specific scenarios, some textual instructions include explicit coordinate information. 
To enhance the generalization ability of the video model, we excluded these coordinate-specific instructions during training.

At the model architecture level, we followed Dynamicrafter~\citep{xing2023dynamicrafter} to modify the text-to-video model of Open-Sora-Plan(T2V)~\citep{pku_yuan_lab_and_tuzhan_ai_etc_2024_10948109} by replacing the first frame and expanding the channel dimensions, enabling the model to take the first frame as a condition. 
This resulted in the Open-Sora-Plan (TI2V) model. 
No structural adjustments were made to other models. 
During training, we preprocessed the data according to each model’s default input format and performed fine-tuning following the official implementation without changing the training settings.
We fine-tuned each model using two different video lengths to enhance the diversity of the video evaluation set: short videos with approximately 20 frames and long videos with around 60 frames, depending on the model’s default training video length. 
The specific lengths are detailed in the Tab.~\ref{tab_training_frames}.

\begin{table}[t]
    \centering
    \footnotesize
    \begin{subtable}{\textwidth}
    \centering
    \begin{tabular}{l|cccccccc}
        \toprule
        \abmc@Acc(\textuparrow) & BC        & FC        & IA        & SA        & VC        & TJ        & EI        & Overall  \\
        \midrule 
        GPT-4o                  & 60.5      & 70.4      & 70.9      & 67.3      & 79.6      & 83.7      & 85.9      & 72.8     \\
        \abevaluator            & \bf 81.2  & \bf 87.5  & \bf 87.5  & \bf 96.4  & \bf 94.5  & \bf 93.8  & \bf 88.8  & \bf 89.4 \\
        \midrule
        GPT-4o@OpenSora         & 60        & 80        & \bf 80    & 50        & 0.0       & \bf 100   & \bf 88.8  & 66.5     \\
        \abevaluator@OpenSora   &\bf  70    & \bf 90    & 60        & \bf 100   & \bf 100   & 22.2      & 80        & \bf 71.6 \\
        \midrule
        GPT-4o@Lavie            & 50        & 66.7      & 75        & 88.8      & \bf 87.5  & \bf 100   & 87.5      & 78.5     \\
        \abevaluator@Lavie      & \bf 80    & \bf 80    & \bf 80    & \bf 100   & \bf 100   & 75        & \bf 100   & \bf 87.9 \\
        \bottomrule
    \end{tabular}
    \end{subtable}
    
    \begin{subtable}{\textwidth}
    \centering
    \begin{tabular}{l|ccccccc}
        \toprule
        \abad@PLCC(\textuparrow)& AE        & IA        & PV        & TJ        & KE        & SF        & Overall   \\
        \midrule
        GPT-4o                  & 0.37      & 0.22      & 0.23      & 0.28      & 0.37      & 0.18      & 0.28      \\
        \abevaluator            &  \bf 0.71 & \bf 0.57  &  \bf 0.50 &  \bf0.58  &  \bf0.65  &  \bf0.58  &  \bf0.60  \\
        \midrule
        GPT-4o@OpenSora         & 0.22      & -0.39     & 0.32      &  \bf0.15  & -0.03     & -0.12     & 0.03      \\
        \abevaluator@OpenSora   &  \bf0.37  & \bf0.55   &  \bf0.34  & 0.06      &  \bf0.28  &  \bf0.41  &  \bf0.34  \\
        \midrule
        GPT-4o@Lavie            & 0.17      & 0.13      & -0.34     & 0.06      & -0.09     & -0.15     & -0.04     \\
        \abevaluator@Lavie      & \bf 0.28  & \bf1.0    &  \bf0.49  &  \bf0.37  & \bf 0.12  &  \bf0.69  & \bf 0.49  \\
        \bottomrule
    \end{tabular}
    \end{subtable}

    \begin{subtable}{\textwidth}
    \centering
    \begin{tabular}{l|cccccccc}
        \toprule
        \abarm@PLCC(\textuparrow)   & AE        & BC        & FC        & IA        & PV        & TJ        & EI        & Overall   \\
        \midrule
        
        GPT-4o                      & 0.07      & 0.18      & 0.20      &  0.32     & -0.14     & -0.01     & -0.14     & 0.07      \\
        \abevaluator                & \bf 0.52  & \bf 0.43  & \bf 0.43  & \bf 0.43  & \bf 0.20  & \bf 0.56  & \bf 0.44  & \bf 0.43  \\
        \midrule
        GPT-4o@OpenSora             & -0.45     & -0.03     & \bf 0.08  & 0.0       & 0.04      & -0.23     & 0.14      & -0.06     \\
        \abevaluator@OpenSora       & \bf 0.25  & \bf 0.35  & 0.05      &  \bf 0.42 & \bf 0.89  & \bf 0.89  & \bf 0.44  & \bf 0.47  \\
        \midrule
        GPT-4o@Lavie                &  0.11     & -0.07     & 0.42      & \bf 0.42  & 0.21      & 0.31      & -0.21     & 0.17      \\
        \abevaluator@Lavie          &\bf0.33    & \bf 0.04  & \bf 0.69  & 0.40      & \bf 0.89  &  \bf 0.67 & \bf 0.06  & \bf 0.44  \\
        \bottomrule
    \end{tabular}
    
    \end{subtable}
    \caption{\textbf{Performance comparison between \evaluator and GPT-4o.} \abevaluator{} indicates \evaluator. The other abbreviations are listed in Tab.~\ref{tab:hierarchical_evaluation_dimension}.}
    \label{tab: comp_eval_gpt}
\end{table}

\subsection{\evaluator{} Traing}
\label{sup:evaluator}
\begin{figure}[t]
\centering
\includegraphics[width=0.8\linewidth]{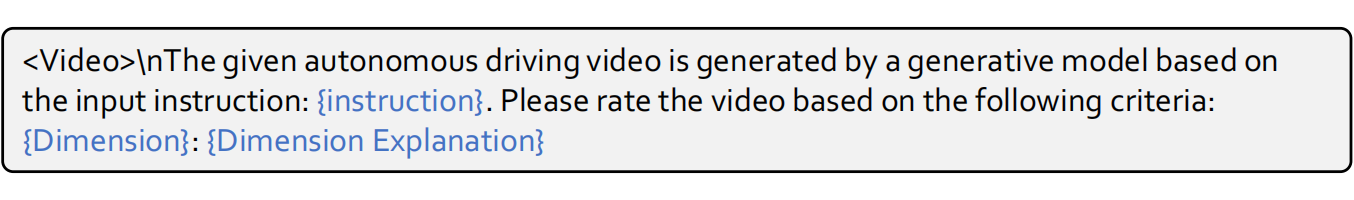}
\caption{\textbf{Prompt template for \ad{}.} The \textcolor[HTML]{4472C4}{\{item\}} is replaced with specific content.} 
\label{fig:prompt_sample}
\end{figure}

The \evaluator{} is trained based on Flash-VStream~\citep{zhang2024flashvstreammemorybasedrealtimeunderstanding}, where only  LoRA~\citep{hu2021loralowrankadaptationlarge} parameters are trained. 
The model's input consists of a sampled video, represented as multiple frames, along with a prompt. The prompt includes the current scenario, the instruction input for video generation, the dimension being evaluated, and the definition of that dimension. An example of such a prompt is illustrated in Fig.~\ref{fig:prompt_sample}, while the details of the explanation are discussed in Section 2.
We don't use the annotated reason during training for CoT of the evaluator, as the reason labeled by different human varies a lot, hard for model to learn.

We maintain consistent training settings in all three scenarios, with a video sampling frequency of 4. The LoRA settings aligned with those in Flash-VStream. We use AdamW as the optimizer, employ cosine decay for the learning rate scheduler. We train for 4 epochs with a learning rate of 2e-5 and a warmup ratio of 0.03. The training is conducted on 4 A100 80 GPUs. To avoid over-fitting to specific prompts or videos generated by particular models, we carefully filter the \dataset{} to ensure balanced distribution across various generation models and evaluation dimensions. 

We prove the effectiveness and generalizability of through comparison with GPT-4o arcoss the three embodied scenarios, under both finetuned and zero-shot setting, as shown in Tab.~\ref{tab: comp_eval_gpt}. After fine-tuning, the \evaluator{} surpasses GPT4-o in aligning with human preferences across all dimensions in every scenario. This is particularly evident in challenging dimensions, \eg, Embodied Interaction and Trajectory in \abarm{}, where GPT4-o shows a negative correlation, while the \evaluator{} exhibits a strong positive correlation. These results demonstrate the its robust performance, making it suitable for \upstream{}. In zero-shot settings, the \evaluator{} also outperforms GPT4-o in nearly all dimensions, further proving our model's aility to understand videos generated by different models.

\section{Detailed Result of \upstream}
\label{Detailed_Result_of_upstream}

\begin{table}[t]
    \centering
    \caption{\textbf{Evaluation results in \abmc{}.} The abbreviations are listed in Tab.~\ref{tab:hierarchical_evaluation_dimension}.}
    \begin{tabular}{@{}l|ccccccc|c@{}}
        \toprule
        Model          & BC  & FC  & IA  & SA  & VC  & TJ   &  EI & Overall \\ 
        \midrule
        Open-Sora-Plan & 1.4 & 1.9 & 1.7 & 1.7 & \bf2.0 & 1.5  & 1.6  & 1.69 \\
        Lavie          & 1.3 & \bf2.0 & 1.7 & 1.7 & \bf2.0 & \bf\bf2.0  & \bf1.8 & 1.79 \\
        ModelScope     & \bf1.9 & \bf2.0 & \bf2.0 & 1.7 & \bf2.0 & \bf2.0  & 1.75  & \bf 1.91\\
        OpenSora       & 1.6 & 1.9 & 1.6 & 1.8 & \bf2.0 & \bf2.0  & 1.6  & 1.79 \\
        AnimateDiff    & 1.3 & 1.3 & 1.2 & 1.7 & 1.4 & 1.38 & 1.55 & 1.40\\
        DynamicCrafter & \bf1.9 & \bf2.0 & 1.5 & \bf2.0 & \bf2.0 & \bf2.0  & 1.45 & 1.84\\
        EasyAnimate    & 1.4 & 1.8 & 1.5 & \bf2.0 & \bf2.0 & 1.22 & 1.45 & 1.62\\ 
        \bottomrule
    \end{tabular}
    \label{tab:detail_mc_performance}
\end{table}

\begin{table}[t]
    \centering
    \caption{\textbf{Evaluation results in \abad{}.} The abbreviations are listed in Tab.~\ref{tab:hierarchical_evaluation_dimension}.}
    \begin{tabular}{@{}l|cccccc|c@{}}
        \toprule
        Model          & AE   & IA  & PV   & TJ  & KE   & SF  & Overall\\ 
        \midrule
        Open-Sora-Plan & 1.6  & \bf5.0 & 1.55 & 1.4 & 1.45 & 3.2 & 2.37\\
        Lavie          & 2.15 & \bf5.0 & 2.2  & 2.8 & 2.1  & \bf5.0 & 3.21\\
        ModelScope     & 2.8  & \bf5.0 & 3.35 & 4.0 & 3.0  & \bf5.0 & 3.86\\
        OpenSora       & \bf3.55 & \bf5.0 & \bf4.4  & \bf4.8 & \bf3.65 & \bf5.0 & \bf 4.40\\
        AnimateDiff    & 1.55 & \bf5.0 & 1.55 & 1.0 & 1.3  & 3.8 & 2.37\\
        DynamicCrafter & 2.6  & 4.0 & 3.4  & 3.8 & 2.65 & 5.0 & 3.57\\
        EasyAnimate    & 1.5  & 3.4 & 1.4  & 1.4 & 1.3  & 2.6 & 1.93\\ 
        \bottomrule
    \end{tabular}
    \label{tab:detail_ad_performance}
\end{table}

\begin{table}[t]
    \centering
    \caption{\textbf{Evaluation results in \abarm{}.} The abbreviations are listed in Tab.~\ref{tab:hierarchical_evaluation_dimension}.}
    \begin{tabular}{@{}l|ccccccc|c@{}}
        \toprule
        Model          & AE   & BC   & FC   & IA   & PV   & TJ  & EI   & Overall \\ 
        \midrule
        Open-Sora-Plan & \bf4.0  & 4.0  & \bf4.0  & 1.0  & 4.9  & \bf5.0 & 4.0 & 3.84 \\
        Lavie          & 3.8  & 3.9  & \bf4.0  & 1.8  & 4.95 & \bf5.0 & 4.1 & 3.94 \\
        ModelScope     & 3.63 & 4.1  & \bf4.0  & 1.18 & 4.9  & \bf5.0 & 4.0 & 3.83 \\
        OpenSora       & 3.85 & 4.0  & 3.95 & 1.3  & 4.75 & \bf5.0 & 4.1 & 3.85 \\
        AnimateDiff    & 3.8  & 3.9  & \bf4.0  & 1.0  & 4.95 & \bf5.0 & 4.1 & 3.82 \\
        DynamicCrafter & 3.97 & \bf4.08 & \bf4.0  & \bf2.6  & \bf5.0  & \bf5.0 & \bf4.31  & 4.14\\
        EasyAnimate    & 3.55 & 3.45 & 3.65 & 1.2  & 4.8  & 4.3 & 3.45 & 3.49\\ 
        \bottomrule
    \end{tabular}
    \label{tab:detail_arm_performance}
\end{table}

Tabs.~\ref{tab:detail_mc_performance}-\ref{tab:detail_arm_performance} present the comprehensive evaluation results for 7 video generation models across three scenarios, including the scores for each dimension and the mean scores representing the overall performance of the models. In \abmc, although our scoring is binary, we display scores on a scale of 1-2 for consistent comparison. In addition to the conclusions mentioned in the main text, we can observe the following findings.

In \abmc, most models achieve high scores in Velocity, largely due to the limited occurrences of object movement in the generated videos. Generating dynamic embodied environments with moving objects presents a significant challenge for current models.  Additionally, the consistency between the generated videos and the scenarios specified in the instructions is higher than the alignment with the task-oriented instructions. This indicates that while the models can generate corresponding scenes, they struggle to reason about the temporal actions necessary for task completion.

In \abad, the quality of the generated videos significantly declines due to the complexity of outdoor driving scenarios. The models must understand and generate various traffic elements, \eg, roads, background buildings, pedestrians, and vehicles, while also producing dynamic content, with each element requiring reasonable speed. This presents substantial challenges. However, top-performing models, \eg, OpenSora, manage to achieve the highest scores across all metrics.

In \abarm, the primary issue lies in Instruction Alignment. The video generation models struggle to comprehend the input instructions and generate appropriate actions to complete the tasks, instead moving aimlessly without clear objectives. This lack of targeted movement reduces potential errors related to object interaction or penetration, resulting in artificially inflated scores in  Embodied Interaction and Trajectory. Current video generation models struggle in effectively addressing robotic manipulation tasks.

\section{\downstream-\abmc}
In this section, we provide additional details about \downstream-\mc ~that are not covered in the main paper due to space limitations.
Minecraft has emerged as a popular open-world environment for developing generalist embodied agents~\citep{lifshitz2024steve,qin2024mp5,zhou2024minedreamer} due to its diverse tasks (e.g., survival, harvesting, crafting, combat, and creative tasks), varied environments, and interactive mobs, all of which require generalized agent capabilities. 
Previous works~\citep{qin2024mp5,wang2023describe,wang2023voyager} have primarily focused on exploring the capabilities of LLMs or MLLMs as \pretextmodel at the $S_1$ stage. 
However, no prior research has conducted closed-loop evaluations of World Simulators at the $S_3$ stage within Minecraft. 
To address this gap, we leverage the Steve-1 pipeline to assess the performance of Video Generation Models as World Simulators in \mc.

\subsection{Detailed Description}

In \downstream-\mc, we adapt the action space of Steve-1~\citep{lifshitz2024steve} to develop a pipeline for the Video Generation Model, enabling it to function as a low-level embodied controller. 
Additionally, we employ Programmatic Evaluation to benchmark the low-level embodied control capabilities of the Video Generation Model as World Simulators. 
These tasks are comprehensive, requiring the combination of multiple atomic actions and smooth scene transitions. 
Each aspect rigorously tests the coherence of the generated content, the consistency with given instructions, and the model's ability to interact effectively with the environment.

\textbf{Testing.}
We evaluated performance in \abmc ~using five tasks: collecting wood, collecting dirt, collecting seeds, exploring the area, and vertical digging. 
To reduce evaluation randomness, we selected the most suitable initialization environments for each task (e.g., the agent is initialized in a forest for the wood collection task).
During testing, for each task, we randomly select one description from various task instructions and input it into the World Simulator to generate the corresponding video. 
The video is then continuously translated into actions by a pre-trained goal-based video-to-action model, which executes until the test time expires. 
Each task runs for 10 trials with distinct environment seeds, with a limit of 3,000 frames (\ie, 2.5 minutes of gameplay).

\textbf{Training.}
Due to the low video quality produced by the open-source video generation model based on the provided instructions, we applied additional fine-tuning using data from the \abmc ~simulation environment. 
For Video Generation Model fine-tuning, we use OpenAI Contractor Gameplay Dataset~\citep{baker2022video} which is the same as \abmc ~in \upstream. The training setting could be found in Sup.~\ref{sup_b2}.
For pre-trained goal-based video-to-action model, we use pre-trained Steve-1(visual) model without extra fine-tuning.

\textbf{Metrics.}
We calculate programmatic evaluation metrics by tracking the MineRL environment state throughout each evaluation episode. Several metrics are measured, including travel distance and early-game item collection. Travel distance is defined as the agent's maximum displacement on the horizontal (X-Z) plane from its initial spawn point. Dig depth is defined as the agent's maximum displacement on the vertical (Y) axis from its initial spawn point. For an early-game inventory, we record the maximum count of logs, seeds, and dirt items observed in the agent's inventory during the episode.

\subsection{Actions}

We use the part of the action space of~\citep{baker2022video} which encompasses nearly all actions available to human players, including keypresses, mouse movements, and clicks. The specific binary actions used in our setup are listed in Tab~\ref{sup_mc_as}.

\begin{table}[htbp]
\centering
\caption{\textbf{Action Space of \abmc.}}
\begin{tabular}{|l|l|}
\toprule
\textbf{Behavior}   & \textbf{Action} \\ \hline
forward   & W key            \\ \hline
back      & S key            \\ \hline
left      & A key            \\ \hline
right     & D key            \\ \hline
jump      & space key        \\ \hline
inventory & E key            \\ \hline
sneak     & shift key        \\ \hline
sprint    & ctrl key         \\ \hline
attack    & left mouse button \\ 
\bottomrule
\end{tabular}
\label{sup_mc_as}
\end{table}

\subsection{Full Result}
Tab.~\ref{sup_full_downstream} presents the evaluation results of several models across five specific tasks (collect wood, collect dirt, collect seeds, travel distance, and dig depth), along with the average (AVG) score for each model. The models are evaluated under two different conditions: Text and Text \& Image. Notably, to ensure that each task falls within a similar score range, we divided the score for the travel distance task by 10 to calculate the AVG score.

\textbf{Performance of Models Under Text Condition.} Open-Sora-Plan and Lavie demonstrate strong performance under the text-only condition, especially in the collect dirt and travel distance tasks. 
Their average scores (26.38 and 26.06, respectively) are very close, indicating consistent and robust performance across tasks.
ModelScope shows an excellent score in the collect dirt task (52.20), but it performs poorly in tasks like collect wood (14.00) and travel distance (240.72), resulting in an overall lower average score (21.050) compared to other text-based models.
OpenSora stands out with the highest overall average score (27.80), excelling particularly in collect dirt (70.20) and travel distance (339.87). This suggests that it is well-adapted to a variety of tasks and exhibits strong task performance.
AnimateDiff shows the weakest performance across all tasks, especially in collect wood (7.40) and collect seeds (3.30), indicating challenges in handling such tasks.

\textbf{Performance of Models Under Text \& Image Condition.} Open-Sora-Plan shows a significant drop in average score under the "Text \& Image" condition, demonstrating that adding image input reduces its performance compared to the text-only condition. 
In particular, its travel distance score drops from 342.91 to 195.14, suggesting that incorporating image data might interfere with certain tasks.
DynamICrafter and EasyAnimate exhibit poor performance across all tasks, especially in collect wood and collect seeds, where they barely complete the tasks (with scores of 0.40 and 0.20, respectively). 
This may indicate a lack of generalization ability in these models when combining image input with text.
Comparing the "Text" and "Text \& Image" conditions, we observe that adding image input does not consistently improve task performance and, in some cases, even degrades it.
We also observed that the success rates of various tasks significantly decrease when an image is added as an additional condition. 
This indicates that the current video generation models need improvement in handling multiple conditional inputs.

\begin{table}[htbp]
    \centering
    \caption{Detail Result of \mc ~in \downstream.}
    \resizebox{\textwidth}{!}{
    \begin{tabular}{@{}p{3cm} >{\centering\arraybackslash}p{2cm}  >{\centering\arraybackslash}p{1cm} >{\centering\arraybackslash}p{2cm} >{\centering\arraybackslash}p{2cm} >{\centering\arraybackslash}p{2cm} >{\centering\arraybackslash}p{2cm} >{\centering\arraybackslash}p{2cm} @{}}
        \toprule
         \multirow{2}{*}{\centering Model}  & \multirow{2}{*}{\centering Condition} & \multirow{2}{*}{\centering AVG} & \multicolumn{5}{c}{Specific Tasks} \\ 
        \cmidrule(lr){4-4}
        \cmidrule(lr){5-5}
        \cmidrule(lr){6-6}
        \cmidrule(lr){7-7}
        \cmidrule(lr){8-8}
         &  &   & Collect Wood & Collect Dirt & Collect Seed&Travel Dis. &Dig Depth  \\
        \midrule
        Open-Sora-Plan & \multirow{5}{*}{\centering Text} & 26.38 & 19.90 & 50.20 & 7.30 &342.91 &20.20  \\
        Lavie &  & 26.06 & 23.50 &56.00 &11.60 &270.20 &12.20  \\
        ModelScope &  & 21.050 & 14.00 &52.20 &6.30 &240.72 &8.70  \\
        OpenSora &  & 27.80 & 21.20 &70.20 &10.40 & 339.87 &3.20  \\
        AnimateDiff &  & 13.10 & 7.40 &22.90  &3.30 &274.19 &4.50  \\
        \midrule
        Open-Sora-Plan& \multirow{3}{*}{\centering Text \& Image} & 10.28 & 11.10 & 12.50 & 2.60 &195.14 &5.70  \\
        DynamiCrafter&  & 4.06 & 0.40 &0.30 &1.30 &130.04 &5.30  \\
        EasyAnimate&  & 4.84 & 0.20 & 0.70 & 1.70 &157.12 &5.90  \\
        \bottomrule
    \end{tabular}
    }
\label{sup_full_downstream}
\end{table}

\subsection{Roll Out}
\begin{figure}[htbp]
    \begin{center}
    \includegraphics[width=1\linewidth]{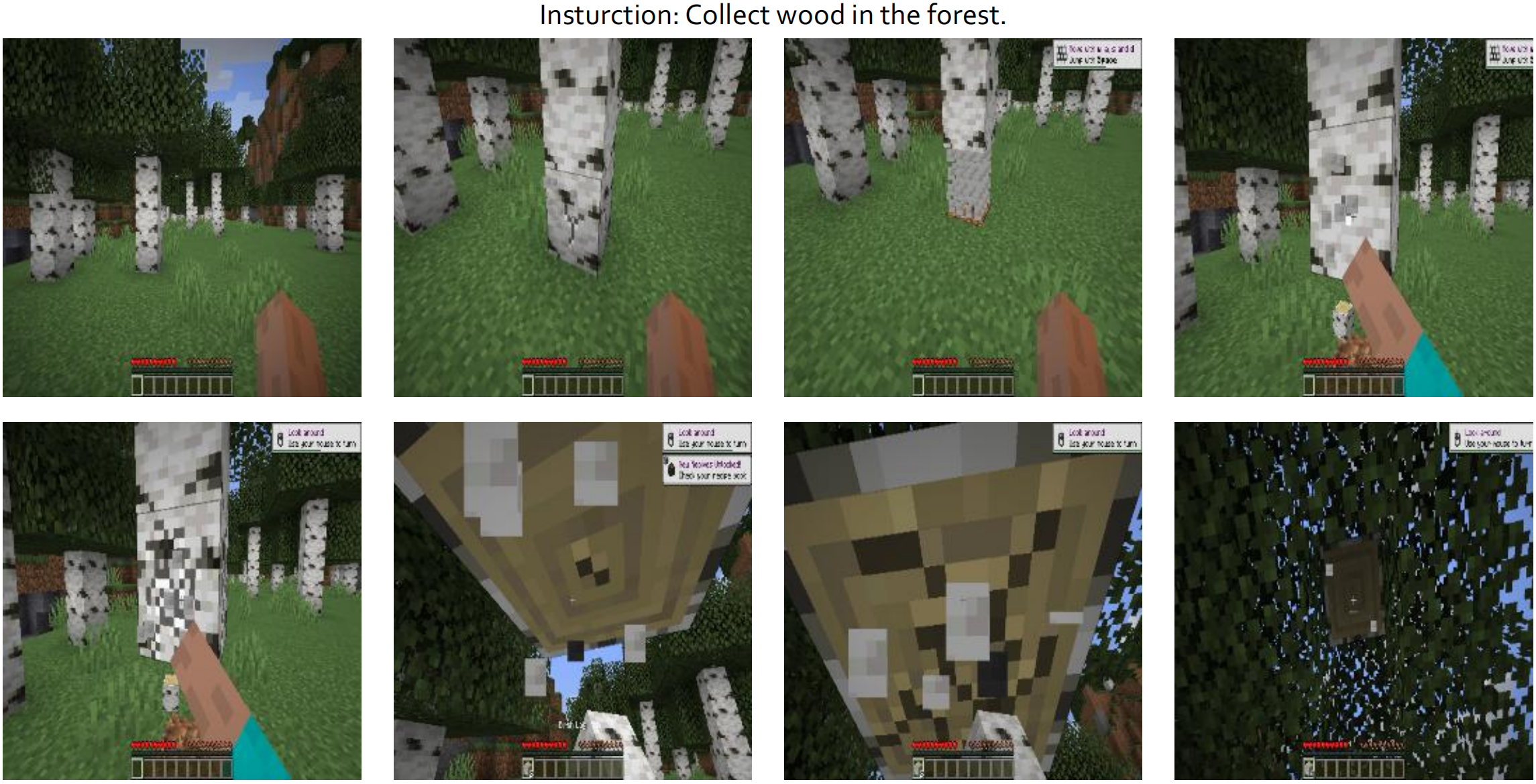}
    \end{center}
    \caption{Rollout of \mc ~in \downstream.} 
    \label{fig:mc_rollout}
\end{figure}

Fig.~\ref{fig:mc_rollout} illustrates the downstream execution process in the \mc, along with the corresponding textual instructions.

\section{\downstream-\abad}
In this section, we provide additional details about \downstream-\ad ~that are not covered in the main paper due to space limitations.

\subsection{Detailed Description}
In \downstream-\ad, we adapt the action space of LMDrive~\citep{shao2024lmdrive} to develop a pipeline for the Video Generation Model, enabling it to function as a low-level embodied controller. 
Additionally, we employ LangAuto (Language-guided Autonomous Driving) CARLA benchmark,  to evaluate the low-level embodied control capabilities of the Video Generation Model as World Simulators. 
These tasks are designed to be comprehensive, spanning all 8 publicly available towns in CARLA, covering a diverse range of scenarios \eg, highways, intersections, and roundabouts. 
Additionally, they account for 16 different environmental conditions, combining 7 distinct weather settings (Clear, Cloudy, Wet, MidRain, WetCloudy, HardRain, SoftRain) with 3 daylight conditions (Night, Noon, Sunset).
Each aspect rigorously tests the coherence of the generated content, the consistency with given instructions, and the model's ability to interact effectively with the environment.

\textbf{Testing.}
We evaluated performance in \ad ~using the LangAuto-Tiny benchmark setting where the route length is shorter than 150 meters. 
We posit that shorter driving distances provide a more effective test of the low-level control capabilities of World Simulators. 
Longer routes typically involve more instructions, which are prone to misalignment with the real-time simulation environment. 
Therefore, we opt to evaluate performance on shorter routes to minimize these discrepancies.
During testing, we randomly select one description from various task instructions and input it into the World Simulator to generate the corresponding video. 
The video is then continuously translated into actions by a pre-trained goal-based video-to-action model, which executes until the test time expires. 
We use the corresponding LangAuto-Tiny instructions and the first-person view rendered by the real-time CARLA simulation environment as input to the video generation model. 
The generated video is then continuously transformed into downstream control signals using a pre-trained video-to-action model until the agent reaches a predefined success zone or the task is terminated due to factors \eg, timeouts or collisions.

\textbf{Training.}
Due to the low video quality produced by the open-source video generation model based on the provided instructions, we applied additional fine-tuning using data from the \abad ~simulation environment. 
For Video Generation Model training, we use LMDrive Training Dataset~\citep{shao2024lmdrive}. 
We preprocessed the training data according to each model’s default input format and performed fine-tuning following the official implementation without changing the training settings.
We fine-tuned each model using a short video generation setting with approximately 20 frames. 
For the video-to-action model, we use pre-trained LMdrive model.
Additional fine-tuning was conducted based on the test requirements. 
We provided the model with arbitrary text instructions and replaced the visual input with the future frame while keeping all other training settings consistent with LMDrive.

\textbf{Metrics.}
We consider eight key metrics introduced by the CARLA Leaderboard~\citep{dosovitskiy2017carla}: Route Completion (RC), Infraction Score (IS), Driving Score (DS), Vehicle Collisions (VC), Pedestrian Collisions (PC), Layout Collisions (LC), Red Light Violations (RV), and Offroad Infractions (OI).
Route Completion refers to the percentage of the total route length that the agent has completed. This metric only accounts for the distance traveled along the predetermined route, where each segment corresponds to a navigation instruction. 
If the agent strays too far from the route, it is considered to have violated the instruction, resulting in the episode being marked as a failure and terminated. 
The Infraction Score tracks any infractions caused by the agent, with penalties applied for collisions or traffic violations through a corresponding discount factor. 
The Driving Score is the product of the route completion ratio and the infraction score, reflecting both driving progress and safety, and is widely regarded as the primary ranking metric.
The precise definitions of the residual metrics can be found in the CARLA documentation~\citep{dosovitskiy2017carla}.

\subsection{Actions}
The video generated by the World Simulator is continuously fed into the video-to-action model to obtain the corresponding waypoints. 
The agent then generates control signals based on the generated waypoints and the conversion strategy used in CARLA.
\subsection{Full Result}
Tab.~\ref{sup_full_downstream_ad} presents the evaluation results of several models across eight metrics.
The evaluation results highlight significant differences in how video generation models perform in autonomous driving tasks.
Open-Sora-Plan stands out in trajectory generation, instruction following, and environment perception, producing high-quality videos that effectively support task execution. 
In contrast, DynamiCrafter and EasyAnimate struggle with generating detailed and consistent video content, particularly when handling complex or dynamic scenes. 
These models require improvements in video generation quality, scene understanding, and task alignment to enhance their performance.

From a video generation perspective, several key areas for future development are identified:
Improved Trajectory Generation: High-quality trajectory generation is essential for accurate control signals. 
Models must focus on generating more coherent and precise trajectories, especially in dynamic environments, to ensure vehicles follow instructions and avoid collisions.
Enhanced Instruction Following: Generated videos should closely align with task instructions, particularly in changing environments, enabling vehicles to adapt quickly while maintaining task accuracy.
Better Environment Perception: Future models need to generate videos that accurately represent complex scenes, \eg, interactions with pedestrians, other vehicles, and varied terrains. 
More detailed and realistic video generation will provide stronger input for real-time decision-making in the control system.

In summary, advancing trajectory accuracy, instruction alignment, and environment representation will be crucial for improving the overall performance of these video generation models in autonomous driving tasks.

\begin{table}[htbp]
    \centering
    \caption{Detail Result of \ad ~in \downstream.}
    \begin{tabular}{@{}lcccccccc@{}}
        \toprule
        Model & DS(\textuparrow) & RC(\textuparrow) & IS(\textuparrow) & VC(\textdownarrow) & PC(\textdownarrow) & LC(\textdownarrow) & RV(\textdownarrow) & OI(\textdownarrow) \\
        \midrule
        Open-Sora-Plan & 31.054 & 38.249 & 0.767 & 2.400 & 0.000 & 4.401 & 1.133 & 3.514 \\
        DynamiCrafter    & 24.491 & 37.189 & 0.599 & 5.030 & 0.000 & 4.896 & 0.937 & 3.221 \\
        EasyAnimate    & 17.414 & 28.475 & 0.607 & 0.000 & 0.000 & 29.344 & 0.000 & 1.690 \\
        \bottomrule
    \end{tabular}
\label{sup_full_downstream_ad}
\end{table}

\subsection{Roll Out}
\begin{figure}[htbp]
    \begin{center}
    \includegraphics[width=1\linewidth]{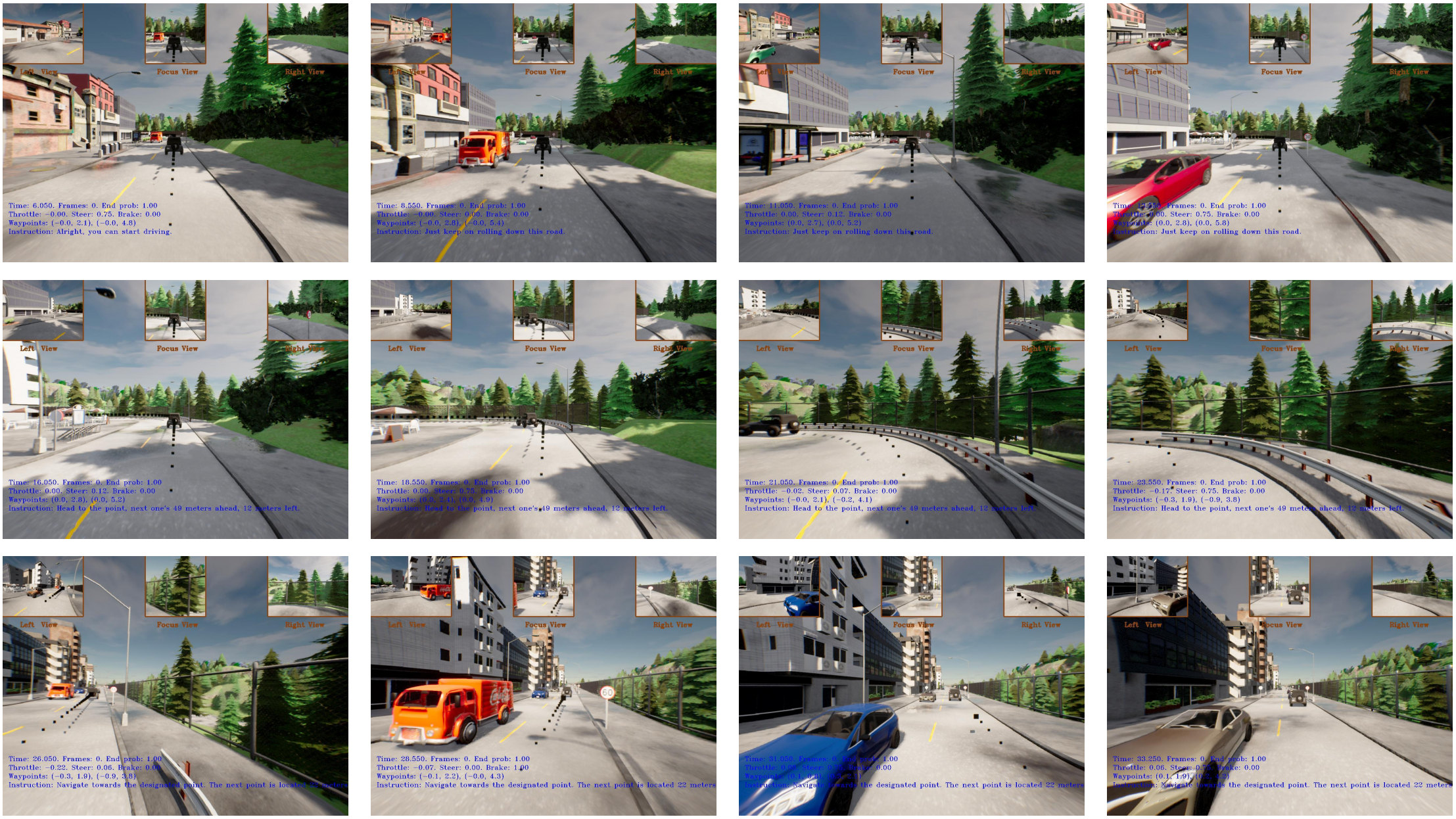}
    \end{center}
    \caption{Rollout of \ad ~in \downstream.} 
    \label{fig:carla_rollout}
\end{figure}

Fig.~\ref{fig:carla_rollout} illustrates the downstream execution process in the \ad, the corresponding text instructions can be found in the lower left corner of each frame.

\section{\downstream-\abarm}
In this section, we provide additional details about \downstream-\arm ~that are not covered in the main paper due to space limitations.

\subsection{Detailed Description}
We primarily conduct our experiments on the CALVIN benchmark~\citep{mees2022calvin}, which is specifically designed for long-horizon, language-conditioned manipulation tasks. 
CALVIN includes four simulated environments (labeled A, B, C, and D) that differ in textures and object placements.
Each environment features a Franka Emika Panda robot positioned next to a desk with various manipulable objects.
The evaluation protocol tests model performance across 1,000 unique instruction chains, each consisting of five distinct tasks. 
By providing an extensive dataset paired with natural language annotations, the CALVIN benchmark can provide a close-loop evaluation platform for evaluating World Simulator to test its generation and generalization capabilities.

\textbf{Testing.}
We evaluated performance in \arm ~using the CALVIN benchmark benchmark, policy models are trained on demonstrations from environments A, B, and C, and evaluated in a zero-shot manner in environment D. 
During the testing phase, we leverage World Simulators and a pre-trained video-to-action model to tackle novel manipulation tasks guided by user-specified natural language commands.
Given a current observation, we generate future video predictions using the World Simulator for the manipulation task with text instruction. 
Once the video is sampled, we then execute the video-to-action policy conditioned on for $k$ timesteps, where $k$ is a testing hyperparameter. After $k$ timesteps, the video prediction is refreshed by sampling from the World Simulator again, and the process is repeated. 

\textbf{Training.}
Due to the low video quality produced by the open-source video generation model based on the provided instructions, we applied additional fine-tuning using data from the \abarm ~simulation environment. 
For Video Generation Model training, we use Calvin(ABC\_D) datset~\citep{mees2022calvin}. 
We preprocessed the training data according to each model’s default input format and performed fine-tuning following the official implementation without changing the training settings.
We fine-tuned each model using a short video generation setting with approximately 20 frames. 
For the video-to-action model, we use a pre-trained Susie policy without extra fine-tuning.

\textbf{Metrics.}
We report the success rates and the average task length completed (out of five tasks) for each evaluation sequence.

\subsection{Actions}
For low-level control, we utilize the same action space as Calvin~\citep{mees2022calvin}.

\subsection{Full Result}
Based on the results shown in Tab.~\ref{tab:CALVIN}, Open-Sora-Plan demonstrates consistent performance, with an average task length of 2.95, indicating its ability to reliably complete task sequences. While DynamiCrafter achieves a higher success rate of 0.95 on the initial task, its performance declines as task complexity increases, suggesting limitations in handling longer manipulation sequences. EasyAnimate, although moderately successful in completing early tasks, experiences a sharp decline in performance as task difficulty rises, reflected in its lower average task length of 2.05.

Overall, the models' ability to consistently complete multiple tasks in succession showcases their potential in downstream applications, with Open-Sora-Plan emerging as the most capable. However, the observed decrease in success rates as task complexity increases highlights the need for further improvements in video-to-action translation, particularly in addressing the challenges posed by longer and more complex manipulation sequences.

\begin{table}[htbp]
\caption{Detail Result of \arm ~in \downstream.}
\centering
\begin{tabular}{lcccccc}
\bottomrule[1pt]
\multirow{2}{*}{Method} &  \multicolumn{5}{c}{Task completed in a row (\%) $\uparrow$ } & \multirow{2}{*}{Avg. Len. $\uparrow$ }  \\
                        &      1 & 2 & 3 & 4 & 5 & \\
\hline
Open-Sora-Plan     &  0.85	&0.70	&0.60	&0.40	&0.40 & 2.95\\  
DynamiCrafter      &  0.95	& 0.75	&0.55	&0.25	&0.25 & 2.75\\ 
EasyAnimate         &  0.90	&0.60	&0.35	&0.10	&0.10 & 2.05\\  
\bottomrule[1pt]
\end{tabular}
\label{tab:CALVIN}
\end{table}

\subsection{Roll Out}
\begin{figure}[htbp]
    \begin{center}
    \includegraphics[width=1\linewidth]{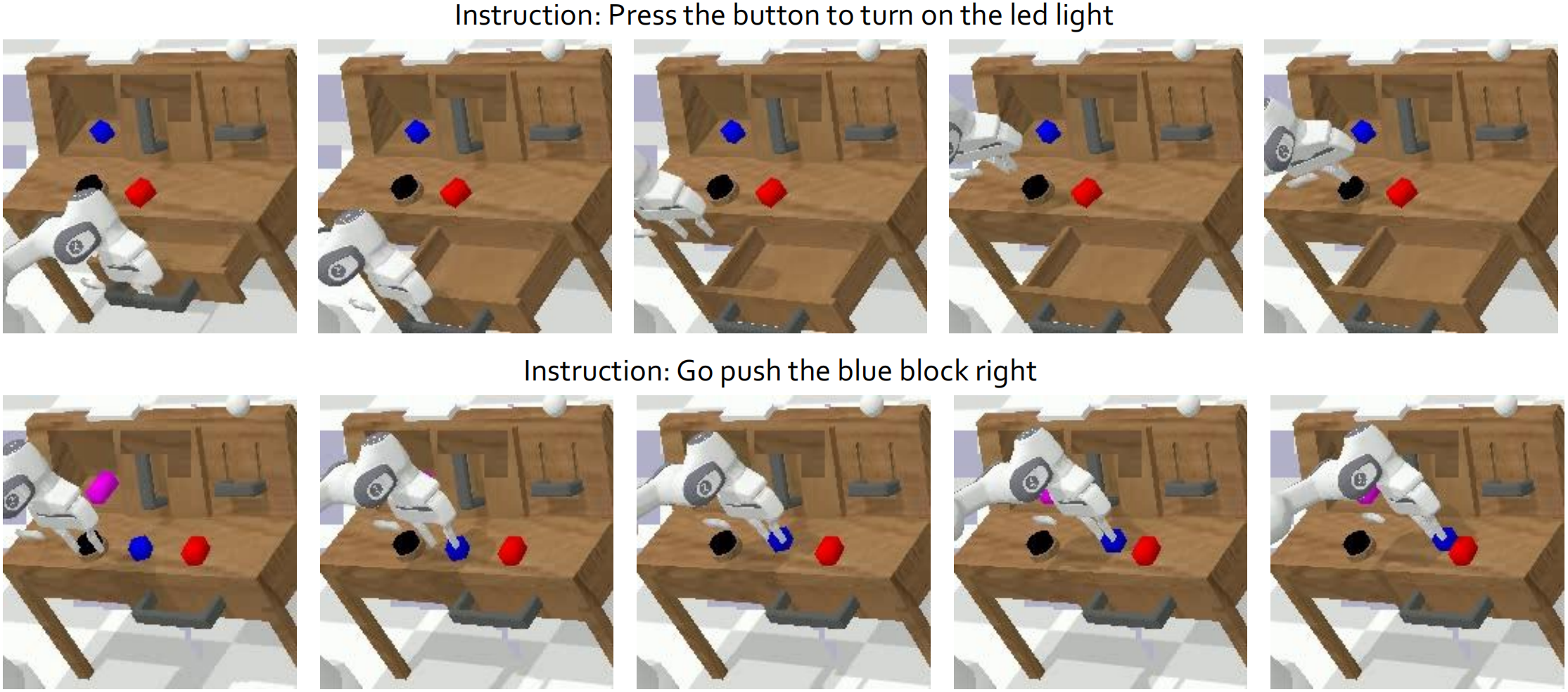}
    \end{center}
    \caption{Rollout of \arm ~in \downstream.} 
    \label{fig:calvin_rollout}
\end{figure}

Fig.~\ref{fig:calvin_rollout} illustrates the downstream execution process in the \arm, along with the corresponding textual instructions.

\end{document}